\documentclass{article}
\usepackage{iclr2027_conference,times}

\usepackage{amsmath,amsfonts,bm}

\def\eqref#1{equation~\ref{#1}}

\def\1{\bm{1}}

\DeclareMathAlphabet{\mathsfit}{\encodingdefault}{\sfdefault}{m}{sl}
\SetMathAlphabet{\mathsfit}{bold}{\encodingdefault}{\sfdefault}{bx}{n}

\usepackage{hyperref}
\hypersetup{colorlinks=true,allcolors=black}
\usepackage{url}
\usepackage{graphicx}
\usepackage{booktabs}
\usepackage{amsmath,amssymb}
\usepackage{microtype}
\usepackage{multirow}
\usepackage{array}
\usepackage{xspace}
\usepackage{xcolor}
\usepackage{tikz}
\usetikzlibrary{arrows.meta,positioning}
\usepackage{listings}

\definecolor{CodeBackground}{HTML}{F7F7F7}
\definecolor{CodeText}{HTML}{202020}
\definecolor{CodeKeyword}{HTML}{0000FF}
\definecolor{CodeFunction}{HTML}{795E26}
\definecolor{CodeVariable}{HTML}{001080}
\definecolor{CodeComment}{HTML}{008000}
\definecolor{CodeString}{HTML}{A31515}
\definecolor{CodeNumber}{HTML}{098658}
\definecolor{CodeRule}{HTML}{D0D0D0}

\lstdefinelanguage{DiffCompressPseudo}{
  morekeywords={for,if,in,return,break,and,or,not},
  morekeywords=[2]{GenerateCandidates,DiagonalL1,LocalDistortion,SnapSignificand,ExactCandidateSubset,CodecBytes,BehaviorDist,sort_by_rate,sort,add,remove,evaluate,Freeze},
  morekeywords=[3]{Input,State},
  sensitive=true,
  morecomment=[l]{\#},
  morestring=[b]",
  morestring=[b]'
}

\title{Neural Compression over Equivalence Classes: Exact Gauges and Functional Rate--Distortion}

\author{Toshiaki Koike-Akino}

\iclrfinalcopy

\begin{document}
\maketitle

\begin{abstract}
Neural networks admit many parameterizations of the same or nearly the same function, but deployment typically stores whichever parameter vector training happens to produce. We show that this representational freedom can itself be used for compression. Rather than redesigning the codec, we search over exact function-preserving transformations and controlled functional perturbations, using the unchanged physical compressor itself to select cheaper representations. Exact gauge families provide a zero-distortion reference under the strict audit; allowing small functional changes exposes much larger rate--distortion gains. Across SmolLM2 and Qwen3, our method reduces Zstandard-compressed checkpoint size by up to 23.4\%, with similar relative reductions under gzip, Brotli, and ZipNN, establishing trained-network representational freedom as a practical storage resource.
\end{abstract}

\section{Introduction}
Training produces one parameter vector for a neural network, but that vector is not the only way to represent the learned computation. Hidden channels can be permuted, neighboring layers can be rescaled together, and internal bases can sometimes be changed without altering the network function. These familiar non-identifiabilities create a practical opportunity: different parameterizations of the same predictor can produce different byte patterns and therefore different compressed sizes.

We study this representational freedom as a compression resource. In the strictest setting, we search only transformations that preserve the neural function exactly. When small behavior changes are acceptable, we enlarge the search to nearby models whose activations or outputs remain within a declared tolerance. This gives a natural progression from exact equivalence to functional rate--distortion (RD), without requiring the stored tensors to match the original checkpoint bit-for-bit.

We keep the physical compressor unchanged, using standard backends such as Zstandard and gzip. Algebraic structure and differentiable signals only propose representations; each is audited for functional admissibility and scored by the real compressor, so measured bytes---not a smooth surrogate---remain the objective.

\begin{figure*}[t]
  \centering
  \includegraphics[width=0.98\textwidth]{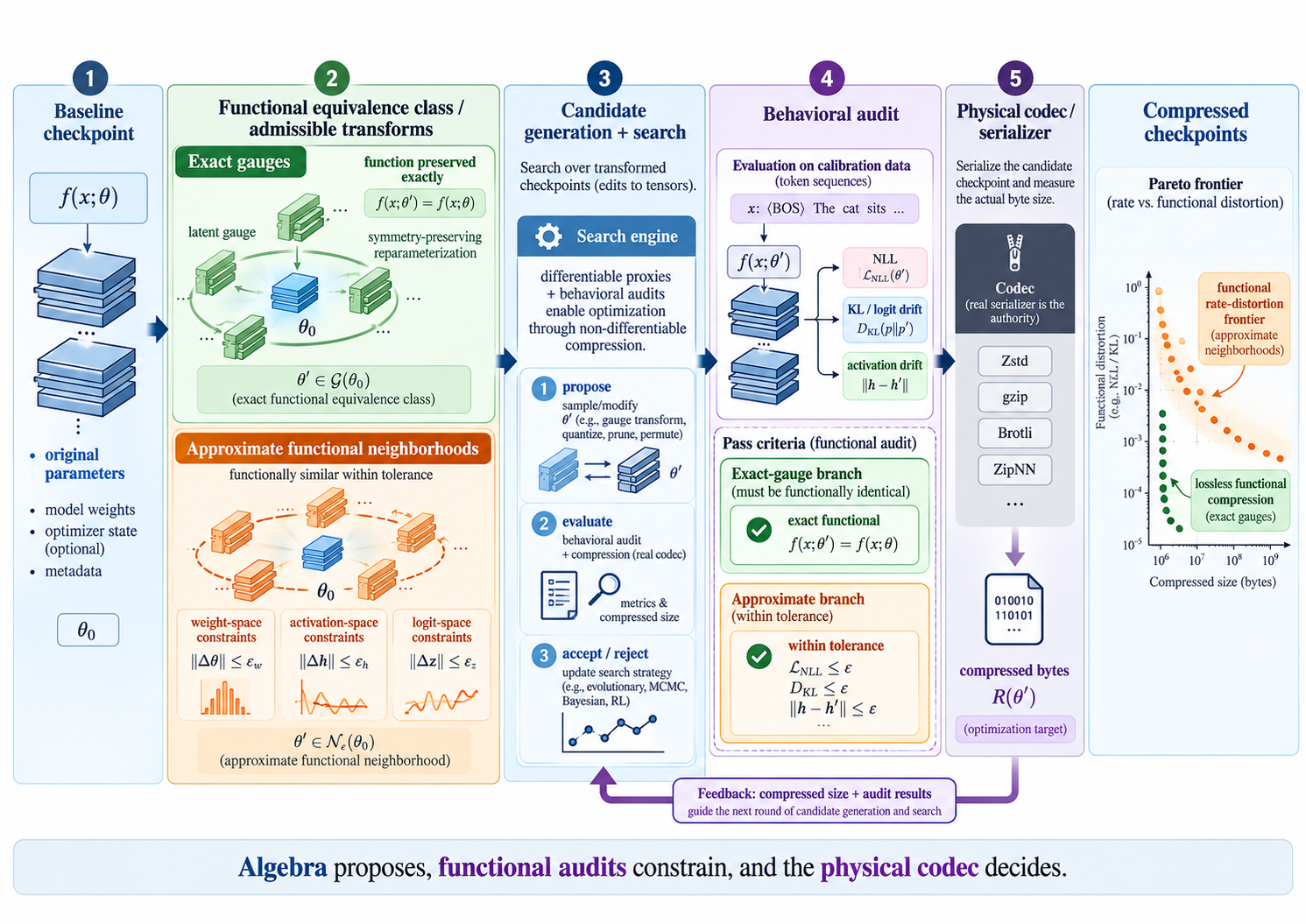}
  \caption{\textbf{Conceptual overview.} Exact gauges move on a function-preserving orbit; controlled neighborhoods enlarge it. Functional audits constrain behavior, and the unchanged physical codec selects the smallest admissible representation.}
  \label{fig:concept}
\end{figure*}

The experiments support three main observations. First, exact function-preserving transformations provide real but comparatively modest compression headroom. Second, controlled functional neighborhoods expose much larger gains: on SmolLM2-360M, the measured Zstandard reduction reaches 23.4\% along the rate--distortion frontier. Third, the effect is not tied to one backend or one architecture: the selected endpoints retain similar relative gains under gzip, Brotli, and ZipNN, and the approach also reduces compressed size on Qwen3. Detailed operating points, preservation audits, scale studies, and negative cases are deferred to the experimental sections and appendices.

The contributions are:
\begin{itemize}
  \item We formulate \textbf{neural compression over functional equivalence classes} as minimizing measured physical storage cost over exact function-preserving representations or explicitly audited functional neighborhoods.
  \item We distinguish an \textbf{exact functional regime} from a \textbf{functional rate--distortion regime}, making clear what is preserved when the parameter bit pattern changes.
  \item We develop a proposal--audit--measure pipeline in which model structure proposes candidates while the unchanged physical compressor remains the final byte-level authority.
  \item We demonstrate the compression hierarchy across scales, architectures, and codecs, with controlled negative cases that delimit useful representational freedom.
\end{itemize}

\section{From Exact Equivalence to Functional Rate--Distortion}
To separate the learned function from its stored representation and storage backend, let $f_\theta$ denote a neural network, $S(\theta)$ a deterministic serialization, and $c$ a concrete physical compressor. The physical cost of a representation is simply the byte count returned by that unchanged backend:
\begin{equation}
C_c(\theta)=\bigl|c(S(\theta))\bigr|.
\end{equation}
Thus the optimization target is not a learned entropy model or differentiable rate proxy, but the realized byte count of the serialized checkpoint.

\subsection{Exact function-preserving compression}
For exact equivalence, we optimize
\begin{equation}
\mathcal{E}_0(\theta)=\{\theta' : f_{\theta'}=f_\theta\},
\qquad
\min_{\theta'\in\mathcal{E}_0(\theta)} C_c(\theta').
\label{eq:exact-class}
\end{equation}
The physical codec is lossless on serialized bytes, while the representation transform preserves the network function. Because the optimized parameterization need not recover the original checkpoint bytes, we call this the exact-functional regime rather than a lossless regime. The function defines an equivalence class on which $C_c$ may vary, so exact compression selects a representative within a fixed function; Appendix~\ref{app:theory} formalizes this quotient-space view.

For the transformer multilayer perceptron (MLP) and attention blocks studied here, exact search covers hidden-channel permutations, reciprocal power-of-two Swish-gated linear unit (SwiGLU) scalings, grouped attention value/output gauges, and low-rank factor gauges. Table~\ref{tab:exact-gauges} summarizes them; Appendix~\ref{app:method} gives derivations and stored-precision audit tiers.

\begin{table}[t]
\caption{Exact function-preserving gauge families. Algebraic identities are audited after storage; strict runs additionally require zero sampled output error.}
\label{tab:exact-gauges}
\centering
\setlength{\tabcolsep}{3pt}
\begin{tabular}{@{}p{0.24\linewidth}p{0.52\linewidth}p{0.20\linewidth}@{}}
\toprule
Gauge family & Representative transform & Preserved object \\
\midrule
\mbox{SwiGLU channel/scale} & $W_g'=PW_g$, $W_u'=DPW_u$, $W_d'=W_dP^{-1}D^{-1}$ & MLP function \\
Attention value/output & $V'=\Gamma^{-1}V$, $O'=O\Gamma$ & value/output map \\
Low-rank factor & $B'=BM$, $A'=M^{-1}A$ & $BA$ update \\
\bottomrule
\end{tabular}
\end{table}

\emph{Toy intuition.} With $A=\left[\begin{smallmatrix}2&4\\4&8\end{smallmatrix}\right]$, $B=\left[\begin{smallmatrix}1/2&1/4\\1&1/2\end{smallmatrix}\right]$, and $M=\operatorname{diag}(2,4)$, the low-rank factor gauge from Table~\ref{tab:exact-gauges}, $A'=M^{-1}A$, $B'=BM$, gives $A'=\left[\begin{smallmatrix}1&2\\1&2\end{smallmatrix}\right]$, $B'=\left[\begin{smallmatrix}1&1\\2&2\end{smallmatrix}\right]$, yet $B'A'=BA$. The eight stored scalars go from six distinct values (empirical symbol entropy $2.5$ bits/value) to $\{1,2\}$ ($1.0$ bit/value). This only illustrates repetition; selection uses measured codec bytes (Appendix~\ref{app:toy-gauge}).

\subsection{What exactness means after finite-precision storage}
The algebraic identities in Table~\ref{tab:exact-gauges} are defined over exact arithmetic, whereas deployment executes stored BF16/FP16/FP32 tensors with finite-precision kernels and potentially reordered reductions. We therefore separate the mathematical gauge from the numerical claim made after serialization. Table~\ref{tab:exactness-tiers-main} summarizes the audit tiers used throughout the paper. The strict branch underlies our \emph{exact functional} terminology. Relaxed stored-precision and data-conditioned branches are reported separately and are never relabeled as bitwise-exact transformations.

\begin{table}[t]
\caption{Exactness tiers. ``Exact gauge'' names the algebraic transform family; the post-storage audit determines the reported claim.}
\label{tab:exactness-tiers-main}
\centering
\setlength{\tabcolsep}{3pt}
\begin{tabular}{@{}p{0.29\linewidth}p{0.36\linewidth}p{0.25\linewidth}@{}}
\toprule
Tier & Post-storage audit & Claim used in this paper \\
\midrule
Strict stored-precision & sampled maximum absolute and relative output errors are both exactly zero & algebraically function-preserving; zero error on the executed audit \\
Finite-precision tolerance & sampled stored-precision error is below a declared numerical threshold & numerically equivalent within tolerance \\
Data-conditioned functional & activation/logit/NLL distortion is measured on real text & functional rate--distortion, not exact equivalence \\
\bottomrule
\end{tabular}
\end{table}

\subsection{Controlled functional neighborhoods}
When exact equality is unnecessarily restrictive, define a behavior distortion $D_{\mathcal D}$ on calibration data $\mathcal D$ and optimize
\begin{equation}
\mathcal{E}_{\epsilon,\mathcal D}(\theta)
=\{\theta':D_{\mathcal D}(f_{\theta'},f_\theta)\le\epsilon\},
\qquad
\min_{\theta'\in\mathcal{E}_{\epsilon,\mathcal D}(\theta)} C_c(\theta').
\label{eq:functional-class}
\end{equation}
For a linear map $W$ with calibration activations $X$, one local criterion is
\begin{equation}
D_{\mathrm{local}}(W,W')
=\frac{\lVert (W'-W)X\rVert_F^2}{\lVert WX\rVert_F^2}
=\frac{\operatorname{tr}[(W'-W)G(W'-W)^\top]}
       {\operatorname{tr}[WGW^\top]},\quad G=XX^\top/T.
\label{eq:local-distortion}
\end{equation}
For $\Delta W=W'-W$, $\mathbb E\|\Delta W x\|_2^2=\operatorname{tr}(\Delta W\Sigma\Delta W^\top)$, so $G=XX^\top/T$ is the empirical plug-in geometry of linear-layer output distortion (Appendix~\ref{app:theory}). In the largest output-functional experiments, the primary distortion is the mean Kullback--Leibler (KL) divergence between original and candidate next-token distributions. Pinsker's and Jensen's inequalities bound the average one-step total variation by $\sqrt{D_{\mathrm{KL}}^{\mathrm{mean}}/2}$; this does not imply rollout identity (Appendix~\ref{app:theory}). Negative log-likelihood (NLL)/perplexity, tail KL, and top-1 agreement are independent diagnostics or held-out audits. Matched pure-KL frontiers receive post hoc codec and downstream audits, with complete WikiText test and streamed C4 audits at 360M; historical guarded endpoints retain separate provenance.

\subsection{Proposal--audit--physical selection}
The physical objective is never replaced by a differentiable proxy. At each coordinate, algebraic structure or a differentiable surrogate generates a small candidate set. A staged audit cheaply removes inadmissible proposals, measures each surviving whole checkpoint with the real codec, and runs expensive block- or logit-level audits only for byte-improving candidates. A candidate is committed only if it satisfies the declared functional condition and strictly lowers the physical objective. Appendix~\ref{app:controlled} shows that differentiable rate surrogates can mis-rank discrete codec outcomes even when exact physical reranking remains reliable.

Algorithm~\ref{lst:physical-search-main} gives the output-functional search. Validation and test data never select updates; they are evaluated only after the endpoint is frozen. Exact-gauge search uses the same physical-selection loop with the distortion predicate replaced by the declared exactness audit.

\begin{lstlisting}[caption={Core proposal, physical scoring, and behavior-audit search for one distortion threshold.},label={lst:physical-search-main}]
Input: original state theta0, codec c, distortion limit eps
State: theta = theta0; R = CodecBytes(theta, c)
for sweep = 1,...,max_sweeps:
    accepted = 0
    for tensor coordinate k:
        P = GenerateCandidates(theta0[k], theta[k], ...)
        improving = []
        for cand in P:
            tc = theta with tensor k replaced by cand
            r = CodecBytes(tc, c)          # real whole-model codec
            if r < R: add (r, cand) to improving
        for (r, cand) in sort_by_rate(improving):
            tc = theta with tensor k replaced by cand
            if BehaviorDist(theta0, tc, calibration) <= eps:
                theta, R = tc, r
                accepted += 1
                break
    if accepted == 0: break
Freeze theta; evaluate validation/test/cross-domain afterward.
\end{lstlisting}
Because each commit strictly lowers a nonnegative integer byte count, updates cannot cycle and finite descent must terminate. An exhaustive no-accept sweep is coordinatewise optimal for the scanned proposals (Appendix~\ref{app:theory}).

\subsection{Single- and multibit weight-space search}
We use two complementary free-weight proposal families: significand snapping and $\ell_1$-regularized sparsification. For a floating tensor $W$ with $p$ significand bits, $Q_b$ rounds each significand to a grid with $b$ fewer low-order bits while preserving tensor shape and dtype. Thus $b$ is a \emph{proposal depth}, not deployment bitwidth: $Q_1$ is the single-bit proposal, while multibit search exposes $Q_b$ for $b\in\{1,2,3,4\}$. Appendix~\ref{app:method} gives a BF16 bit-level example.

The second family minimizes a weighted quadratic reconstruction term plus the entrywise $\ell_1$ penalty $\|W'\|_1=\sum_{ij}|W'_{ij}|$, yielding soft thresholding and exact zeros; $L_s(W_0)$ denotes the resulting candidate tuned to target sparsity $s$. The candidate set is a \emph{union}, not a composition,
\begin{equation}
\mathcal P(W_0)=\{Q_b(W_0):b\in\mathcal B\}\;\cup\;\{L_s(W_0):s\in\mathcal S\},
\label{eq:proposal-union}
\end{equation}
where $W_0$ is the immutable original tensor. The dense experiments reported here use $\mathcal S=\{5\times10^{-4},10^{-3},2\times10^{-3},5\times10^{-3}\}$. We call $\mathcal B=\{1\}$ \textbf{1-bit+$\ell_1$} and $\mathcal B=\{1,2,3,4\}$ \textbf{multibit+$\ell_1$}. Every alternative is constructed from $W_0$; later sweeps may replace a candidate but never recursively round an already modified tensor. Appendix~\ref{app:method} gives exact constructions and pseudocode.

\section{Experimental Setup}
We study SmolLM2-135M and SmolLM2-360M causal language models \citep{allal2025smollm2}, with Qwen3-0.6B-Base \citep{yang2025qwen3} as an architecture-transfer target. Whole-model experiments use deterministic sorted-key serialization. Zstandard is the primary optimization backend because it combines strong baseline compression with evaluation that is fast enough for repeated search; gzip and Brotli provide generic cross-codec controls, while ZipNN 0.5.4 in Huffman mode \citep{hershcovitch2025zipnn} provides a tensor-specialized post hoc control. Codec version is part of the experimental condition, and absolute byte counts from different versions are not mixed. We evaluate strict and finite-precision gauges, token-free weight-space perturbations, activation-aware neighborhoods, and output-space functional neighborhoods.

Search and candidate acceptance use only a WikiText-2 raw training calibration sample \citep{merity2017pointer}. The disjoint validation sample never selects updates; after endpoint freeze, its measured distortion defines the empirical RD coordinate in Figure~\ref{fig:llm-rd-scale}. Where reported, WikiText test and streamed C4 (Colossal Clean Crawled Corpus) \citep{raffel2020t5} provide subsequent in-domain and cross-domain audits. We make no universal claim for unseen sequences, instead reporting distortion on increasingly independent data; Appendix~\ref{app:experiments} gives the full protocol, search budgets, cache semantics, and split details.

\section{Large Language Model (LLM) Experiments and Results}
\subsection{Exact gauges establish the exact functional regime}
The strict gauge result anchors Table~\ref{tab:main-hierarchy}. On SmolLM2-135M, whole-model strict gauge search saves 14,416 bytes under Zstandard. The gain is small relative to the approximate regimes, but it anchors the zero-distortion reference: compressed size changes under an algebraically exact gauge while the strict stored-precision audit reports zero output error. Module-framed optimization initially suggested a 67,971-byte gain, but only 6,194 bytes survived monolithic serialization; moving the whole-model codec into the optimization loop recovered 14,416 bytes. Thus even exact gauge optimization must respect the global physical stream rather than add local byte improvements independently.

A public Qwen3 module probe isolates this mechanism more directly. On a reversible channel-major serialization of a contiguous 32-channel SwiGLU window, all eight tested random shared channel permutations preserve the module function yet reduce exact Zstandard size by 18--92 bytes. Eight DiffZstd-guided proposals also reduce the real codec size, but their surrogate/exact Spearman correlation is $-0.193$; the best surrogate-guided candidate saves 78 bytes, below the 92-byte best random proposal under the same evaluation budget. This separates two claims: exact symmetry can change physical compressed size, but a differentiable rate surrogate need not rank discrete codec outcomes reliably; the production codec remains authoritative.

\begin{table}[t]
\caption{Cross-codec transfer of Zstandard-selected SmolLM2-135M representations. Rows are matched to their recorded Zstandard conditions; gzip, Brotli, and ZipNN are post hoc, and multibit+$\ell_1$ uses the matched pure-KL frontier.}
\label{tab:main-hierarchy}
\centering
\begin{tabular}{lrrrr}
\toprule
Representation class & Zstd gain & gzip gain & Brotli gain & ZipNN gain \\
\midrule
Strict exact gauge & 14.4 KB & 6.1 KB & 5.0 KB & 0.3 KB \\
Finite-precision gauge & 16.2 KB & 13.5 KB & 8.8 KB & 1.4 KB \\
Weight-space & 32.5 KB & 25.0 KB & 36.7 KB & -5.0 KB \\
Block activation & 253.7 KB & 225.2 KB & 271.4 KB & 261.4 KB \\
Local activation & 1.43 MB & 1.24 MB & 1.48 MB & 1.44 MB \\
Output functional, 1-bit+$\ell_1$ & 12.84 MB & 10.18 MB & 12.57 MB & 11.77 MB \\
Output functional, multibit+$\ell_1$ & 43.41 MB & 44.20 MB & 44.47 MB & 39.38 MB \\
\bottomrule
\end{tabular}
\end{table}

\subsection{Functional neighborhoods expose much larger physical headroom}
The hierarchy widens dramatically once exact equality is relaxed. Block and local activation save 253,728 and 1,432,379 bytes under pinned Zstandard v1.5.5. The initial logit-constrained sweep also imposed a hard local-activation prefilter, limiting the admissible search. Removing it while retaining output-space behavior checks changes the scale: the 1-bit+$\ell_1$ search edits all 90 eligible MLP projections and reduces the 208,502,655-byte baseline by 12,841,952 bytes (6.1591\%).

The endpoint remains close to the original model on independent data. Table~\ref{tab:llm-heldout-main} summarizes the held-out audit detailed in Appendix~\ref{app:results}. On 8,192 WikiText test tokens, PPL increases by 0.0346\%, mean KL is 0.001856, and top-1 agreement is 97.31\%; on a separately streamed 8,192-token C4 validation sample, these are 0.0624\%, 0.002029, and 97.05\%. Post hoc ZipNN 0.5.4 compresses the original and Zstandard-selected endpoint to 179,380,579 and 167,612,739 bytes, a gain of 11,767,840 bytes, and never selects candidates.

\subsection{Post hoc downstream preservation}
Frozen output-functional endpoints are also evaluated post hoc with EleutherAI \texttt{lm-eval} on tasks never used in search. Table~\ref{tab:downstream-main} reports representative 1-bit+$\ell_1$ endpoints on LAMBADA \citep{paperno2016lambada}, PIQA \citep{bisk2020piqa}, and ARC-Easy \citep{clark2018arc}; Appendix~\ref{app:downstream} adds HellaSwag \citep{zellers2019hellaswag}, ARC-Challenge, normalized variants, and both matched pure-KL frontiers. At 135M, the $8\times10^{-3}$ control reduces Zstandard size by 20.695\% with raw-accuracy deviations within 0.52 percentage points and LAMBADA PPL $+0.35\%$; $10^{-2}$ adds only 0.126 percentage points of compression but raises LAMBADA PPL to $+3.92\%$. The 360M sweep shows the same saturation: $8\times10^{-3}$ reaches 23.344\% with WikiText/C4 PPL $+0.668/+0.293\%$, maximum raw-accuracy deviation 0.49 percentage points, and LAMBADA PPL $+2.35\%$; $10^{-2}$ adds only 0.061 percentage points of compression but raises LAMBADA PPL to $+5.02\%$.

\begin{table*}[t]
\caption{Representative frozen 1-bit+$\ell_1$ SmolLM2 downstream preservation (original $\rightarrow$ compressed). Appendix~\ref{app:downstream} reports full five-task audits and both matched pure-KL multibit+$\ell_1$ frontiers.}
\label{tab:downstream-main}
\centering
\setlength{\tabcolsep}{4pt}
\begin{tabular}{@{}lccc|ccc@{}}
\toprule
& \multicolumn{3}{c|}{SmolLM2-135M} & \multicolumn{3}{c}{SmolLM2-360M} \\
Metric & Original & Endpoint & $\Delta$ & Original & Endpoint & $\Delta$ \\
\midrule
LAMBADA accuracy & 0.4287 & 0.4306 & +0.0019 & 0.5333 & 0.5337 & +0.0004 \\
LAMBADA perplexity & 19.0560 & 19.0044 & -0.0516 & 9.3938 & 9.3794 & -0.0144 \\
PIQA accuracy & 0.6812 & 0.6866 & +0.0054 & 0.7165 & 0.7187 & +0.0022 \\
ARC-Easy accuracy & 0.6427 & 0.6410 & -0.0017 & 0.7033 & 0.7058 & +0.0025 \\
\bottomrule
\end{tabular}
\end{table*}

\begin{table}[t]
\caption{Held-out audits of the frozen SmolLM2-135M 1-bit+$\ell_1$ endpoint (WikiText in-domain; C4 cross-domain).}
\label{tab:llm-heldout-main}
\centering
\begin{tabular}{lrrrrr}
\toprule
Audit & rel. $\Delta$PPL & mean KL & P99 KL & P99.9 KL & top-1 \\
\midrule
WikiText test & +0.0346\% & 0.00185644 & 0.00868125 & 0.0552182 & 97.31\% \\
C4 validation & +0.0624\% & 0.00202928 & 0.00845627 & 0.0911304 & 97.05\% \\
\bottomrule
\end{tabular}
\end{table}

\begin{figure*}[t]
  \centering
  \includegraphics[width=0.98\textwidth]{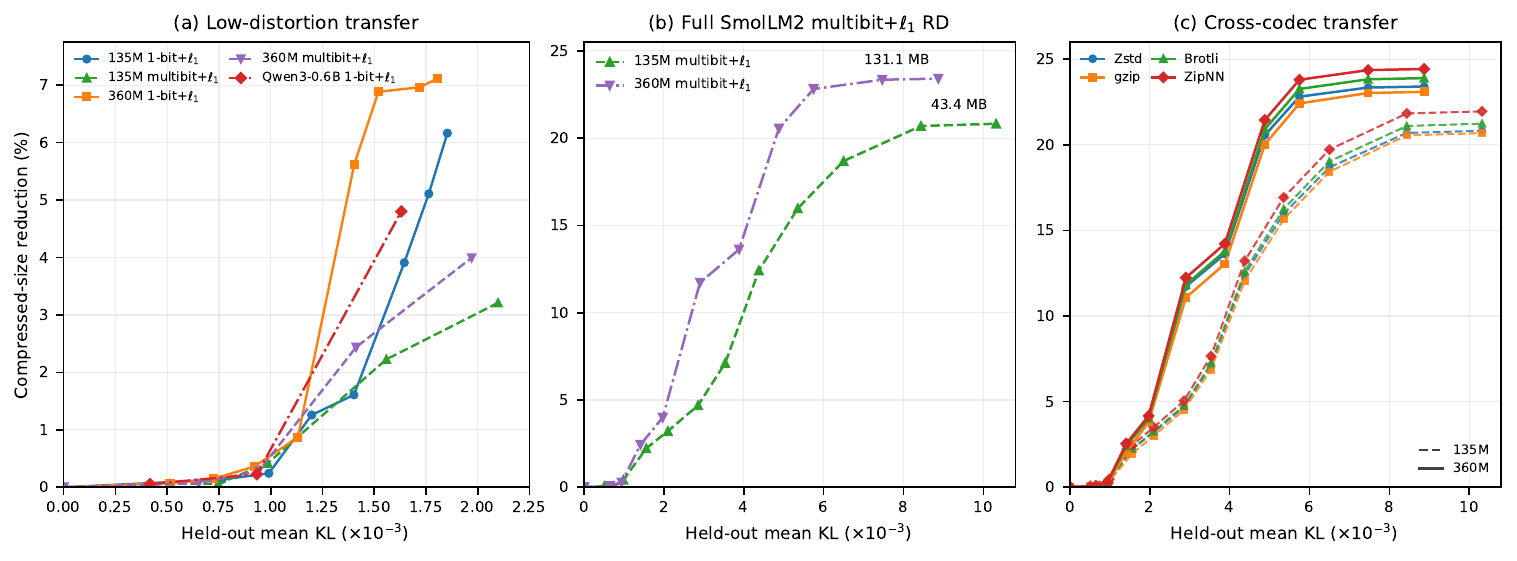}
  \caption{\textbf{Measured LLM functional rate--distortion and codec transfer.} All panels use held-out validation mean token KL on the x-axis. (a) Low-distortion SmolLM2-135M/360M and Qwen3-0.6B. (b) Matched pure-KL SmolLM2 multibit+$\ell_1$ under Zstandard. (c) Frozen 135M/360M endpoints recompressed by four codecs and normalized to each codec baseline; only Zstandard selects.}
  \label{fig:llm-rd-scale}
\end{figure*}

Figure~\ref{fig:llm-rd-scale} combines the corrected SmolLM2-135M and completed SmolLM2-360M multibit+$\ell_1$ pure-KL frontiers, Qwen3-0.6B architecture transfer, and full cross-codec transfer at both SmolLM2 sizes. Both SmolLM2 multibit sweeps use independent restarts and calibration mean-KL-only acceptance.

Panel~(a) shows that the larger multibit+$\ell_1$ candidate set does not uniformly dominate the 1-bit+$\ell_1$ greedy search in the low-distortion regime at either SmolLM2 scale, even though it contains the one-bit significand proposal plus the same $\ell_1$ alternatives. This is consistent with optimization-path dependence across independently restarted trajectories. Panel~(b) shows the additional high-distortion headroom under a matched pure-KL protocol. At 135M, the first multibit+$\ell_1$ endpoint above the frozen 1-bit+$\ell_1$ ceiling occurs at the $3\times10^{-3}$ search control, saving 14.84~MB (7.115\%) at measured validation mean KL 0.003536; the sweep reaches 43.41~MB (20.82\%) at KL 0.010321. At 360M, the same $3\times10^{-3}$ control first exceeds the 39.84-MB 1-bit+$\ell_1$ ceiling, saving 65.65~MB (11.72\%) at measured KL 0.002904, and the sweep reaches 131.09~MB (23.41\%) at KL 0.008871. The widest boundary points have validation $\Delta$NLL $+0.01203$ and $+0.00783$, so we treat them as RD boundary observations rather than low-distortion operating points. The requested calibration KL limits remain search controls; plotted distortion is remeasured on the disjoint validation split after endpoint freeze.

\subsection{Architecture transfer and proposal-family ceilings}
On Qwen3-0.6B, the $2\times10^{-3}$ 1-bit+$\ell_1$ search edits all 84 MLP projections with $Q_1$, saving 44.29~MB (4.802\%) at validation KL 0.001631 and $\Delta$NLL $+0.001082$. Restarts through $8\times10^{-3}$ reproduce the same edit sequence and a zero-accept second sweep, exposing a proposal-family ceiling.

SmolLM2-360M shows the same 1-bit+$\ell_1$ saturation. At $2\times10^{-3}$ it saves 39.84~MB (7.113\%) from a 560.07-MB Zstandard v1.5.7 baseline; a permissive KL-only probe edits all 96 searchable tensors, reaches the same gain, then has a zero-accept sweep, making 39.84~MB a proposal-family ceiling. Multibit+$\ell_1$ reaches 65.65~MB (11.72\%) at validation KL 0.002904 and 131.09~MB (23.41\%) at KL 0.008871. Widest matched pure-KL boundaries reduce Zstandard size by 20.82\% (135M) and 23.41\% (360M); two sizes do not establish a scaling law. Experiment~109 audits all 360M points on WikiText test, streamed C4, and A40-matched downstream tasks: near-saturated points stay close while the widest boundary degrades more sharply.

Table~\ref{tab:llm-scale-main} gives an older Zstandard v1.5.5 scale control across stricter representation classes: exact-gauge savings stay nearly constant, whereas local-activation savings grow from 1.43 to 8.06~MB and the corresponding ZipNN gain grows from 1.44 to 7.26~MB.

\begin{table*}[t]
\caption{Matched SmolLM2-135M/360M scale study under Zstandard v1.5.5 and post hoc ZipNN 0.5.4; positive ZipNN gain indicates transfer.}
\label{tab:llm-scale-main}
\centering
\footnotesize
\setlength{\tabcolsep}{3.5pt}
\begin{tabular}{lrrrrrr}
\toprule
Representation & 135M Zstd & 360M Zstd & 135M ZipNN & 360M ZipNN & 360M mean KL & 360M top-1 \\
\midrule
Strict exact gauge & 14,416 & 14,205 & 256 & 7,742 & -- & -- \\
Weight-space & 32,471 & 90,068 & -4,963 & -25,366 & 0.001302 & 97.72\% \\
Block activation & 253,728 & 397,107 & 261,398 & 363,229 & 0.001355 & 97.71\% \\
Local activation & 1,432,379 & 8,064,543 & 1,437,369 & 7,262,940 & 0.001518 & 97.72\% \\
\bottomrule
\end{tabular}
\end{table*}

\subsection{Cross-codec transfer}
At 135M, Zstandard-selected block-activation, local-activation, and 1-bit+$\ell_1$ endpoints save 0.261/1.437/11.768~MB under ZipNN; matched 360M block- and local-activation endpoints save 0.363/7.263~MB. Weight-space perturbations instead make ZipNN output larger.

Panel~(c) of Figure~\ref{fig:llm-rd-scale} shows matched transfer of both pure-KL frontiers. At the widest boundaries, Zstandard/gzip/Brotli/ZipNN reductions are 20.82/20.68/21.24/21.95\% (135M) and 23.41/23.09/23.89/24.43\% (360M); ZipNN bitwise verifies all 272/290 floating tensors without fallback.

A separate whole-file safetensors check confirms that the effect survives ordinary checkpoint serialization. The 135M raw files remain 269.06~MB before compression (optimized files differ by only 32~B); opaque whole-file Zstandard/gzip/Brotli yield 6.17/4.78/5.90\% reductions for the 1-bit+$\ell_1$ endpoint and 20.81/20.63/21.10\% for the historical guarded multibit+$\ell_1$ boundary. Appendix~\ref{app:whole-safetensors} gives exact sizes; Table~\ref{tab:zipnn-transfer} reports tensor-aware ZipNN separately. Together, these results show that representation shaping complements independent compressors rather than replacing codec design.

\subsection{Mechanism breadth beyond whole-model Zstandard}
Whole-model LLM experiments are the primary evidence, but the proposal--audit--physical-selection architecture was also validated across distinct non-differentiable storage mechanisms. Table~\ref{tab:controlled-systems-main} summarizes controlled studies from Appendix~\ref{app:controlled}; these small or synthetic tests establish mechanism breadth rather than model-scale claims.

\begin{table*}[t]
\caption{Representative mechanism-validation studies across physical storage objectives.}
\label{tab:controlled-systems-main}
\centering
\setlength{\tabcolsep}{3pt}
\begin{tabular}{@{}>{\raggedright\arraybackslash}p{0.12\textwidth}>{\raggedright\arraybackslash}p{0.20\textwidth}>{\raggedright\arraybackslash}p{0.35\textwidth}>{\raggedright\arraybackslash}p{0.21\textwidth}@{}}
\toprule
Physical objective & Admissible freedom / proposal model & Representative measured result & Why it matters \\
\midrule
DEFLATE / gzip & soft Bellman parse + exact byte repair; exact hidden permutation & raw DEFLATE 208$\to$168 bits; gzip 352$\to$312 bits; seed-7 surrogate Pearson 0.93587 & differentiable rate proposal, real codec authority \\
Checkpoint delta & exact hidden permutation + SoftDelta placement & compression-aware alignment evaluated by exact decodable DiffDelta and Git binary patch & representation choice affects cross-checkpoint rate \\
Rabin dedup & exact hidden permutation + SoftDedup & exact trajectory 4,084$\to$3,662 B; oracle recovered; max function error $9.54\times10^{-7}$ & extends beyond entropy/LZ coding \\
Zstandard & structured SoftZstd parser/rate model & exact Zstd 344$\to$216 bits; Pearson 0.98916, Spearman 1.0 & codec-specific proxy can guide proposals \\
Brotli & soft LZ + context/dictionary proxy & differentiable proposal model with unchanged production Brotli final evaluation & same architecture with a different codec grammar \\
\bottomrule
\end{tabular}
\end{table*}

\section{Discussion and Limitations}
The results suggest three regimes. First, exact gauges show that functional equivalence does not imply physical equivalence, although trained models may already occupy stable gauges that limit exact savings. Second, activation-aware neighborhoods expose substantially more freedom than parameter-space distance alone. Third, output-functional neighborhoods can expose an order of magnitude more physical headroom, but necessarily as rate--distortion rather than exact equivalence. The negative cases in Table~\ref{tab:negative-main} reinforce this boundary: natural LoRA trajectories can be nearly gauge-stable even though the same machinery removes large injected gauge drift (Appendix~\ref{app:negative}).

\begin{table}[t]
\caption{Exact-gauge boundary cases: large algebraic freedom need not yield large physical headroom in natural training.}
\label{tab:negative-main}
\centering
\setlength{\tabcolsep}{3pt}
\begin{tabular}{@{}p{0.31\linewidth}p{0.30\linewidth}p{0.33\linewidth}@{}}
\toprule
Case & Exact freedom & Physical outcome \\
\midrule
Static rank-8 Qwen3 LoRA & rank-space $GL(8)$ & best 857 B of 4.29 MB ($0.020\%$) \\
Qwentaur, 5 checkpoints & 784 exact gauge coordinates & 97 B of 60.57 MB ($0.000160\%$) \\
Synthetic scrambled history & full $GL(4)$ & 3,418 B of 6,162 B removed ($55.5\%$) \\
\bottomrule
\end{tabular}
\end{table}

The study does not establish a universally best distortion metric. Mean KL is natural for preserving output distributions, while NLL/perplexity is task- and corpus-dependent. We therefore report multiple diagnostics and separate calibration, validation, test, cross-domain, downstream, and cross-codec audits. The plotted 135M and 360M multibit+$\ell_1$ sweeps use matched pure-KL acceptance; both have complete five-task and four-codec frontier audits, while 360M additionally has complete WikiText test and streamed C4 frontier audits. Historical guarded 135M checkpoints remain only for provenance and the earlier whole-file serialization audit. At both scales, byte savings nearly saturate before LAMBADA perplexity degrades sharply, reinforcing the interpretation of the widest endpoints as RD boundaries rather than preferred preservation points. These audits remain finite: WikiText/C4 each use 8,192 tokens and downstream coverage is five zero-shot tasks, so they do not establish universal preservation over unseen corpora or long-horizon generation.

\section{Related Work}
\textbf{Symmetries and reparameterization.}
Neural networks possess non-identifiabilities under permutations and rescalings; these symmetries can alter parameter-space geometry without changing the represented function \citep{dinh2017sharp}. Permutation-aware alignment has been used for mode connectivity and model merging \citep{entezari2022permutation,ainsworth2023gitrebasin}, and transformer permutation equivariance has been studied directly \citep{xu2024permutation}. Closest to our compression motivation, \citet{martinez2021permute} use function-preserving permutations to facilitate product quantization. We instead optimize the measured output of an arbitrary physical backend directly, include non-permutation gauges, and extend exact invariance to explicitly audited functional neighborhoods.

\textbf{Activation-aware model compression.}
Modern pruning, quantization, and low-rank methods exploit activation information to preserve model behavior under constrained representations \citep{frantar2023gptq,frantar2023sparsegpt,sun2024wanda,dettmers2022llmint8,lin2024awq,yuan2023asvd,wang2025svdllm}. We share the activation-aware principle but optimize a different variable: tensor shapes and dtypes are retained, and candidate weights are selected by the exact output of an unchanged physical compressor rather than by a prescribed bitwidth, sparsity, or rank target.

\textbf{Lossless neural-weight compression.}
ZipNN specializes lossless compression to neural tensors \citep{hershcovitch2025zipnn}. Our approach is complementary: instead of designing a new codec for a fixed model, we optimize which functionally admissible representation is presented to the codec. Cross-codec transfer therefore tests whether a discovered representation is broadly compressible or merely exploits one backend. Appendix~\ref{app:related} gives a fuller comparison.

\section{Conclusion}
Neural compression need not change precision, sparsity, architecture, or codec: trained networks already contain representational freedom. Exact gauges preserve the predictor while changing its physical representation; functional neighborhoods extend this to rate--distortion. Across SmolLM2 scales, headroom grows from KB-scale exact-gauge gains to MB-scale activation-aware and output-functional gains that transfer across generic and tensor-specialized codecs.

The broader lesson is not that every equivalence class offers useful compression headroom: natural LoRA histories provide a clear counterexample. Rather, functional non-identifiability is itself a deployable resource whose value depends on the model, representation family, and physical objective. The proposal--audit--physical-selection framework provides a way to exploit that resource while leaving the final discrete storage system unchanged and authoritative.

\subsection*{AI use statement}
OpenAI ChatGPT assisted with software prototyping and refactoring, test generation, experiment scripting and job-launcher preparation, literature triage, and manuscript drafting and editing. Figure~\ref{fig:concept} was generated with OpenAI ChatGPT's image-generation capability and selected and reviewed by the authors. All experiments were conducted by the authors, primarily on NVIDIA A40 GPUs. AI-assisted code was validated with automated tests and consistency checks, and AI-assisted citation suggestions were checked against primary publication records. OpenAI ChatGPT did not create, alter, or impute experimental measurements; empirical plots, tables, and quantitative claims derive from author-executed code and recorded artifacts. The authors take responsibility for the final manuscript and evidence.

\bibliography{references}
\bibliographystyle{iclr2027_conference}

\appendix
\section{Mathematical Details of the Representation Classes}
\label{app:method}

\subsection{Physical objective and functional neighborhoods}
Let $S(\theta)$ serialize all stored tensors in a fixed deterministic order and let $c$ be a concrete physical codec implementation. We optimize the measured byte length
\begin{equation}
C_c(\theta)=|c(S(\theta))|.
\end{equation}
The exact equivalence class is
\begin{equation}
\mathcal E_0(\theta)=\{\theta':f_{\theta'}(x)=f_\theta(x)\ \forall x\}.
\end{equation}
For finite-precision or data-conditioned equivalence, let $D_{\mathcal D}$ be a behavior distortion measured on samples from $\mathcal D$ and define
\begin{equation}
\mathcal E_{\epsilon,\mathcal D}(\theta)=
\{\theta':D_{\mathcal D}(f_{\theta'},f_\theta)\le\epsilon\}.
\end{equation}
Our generic problem is therefore
\begin{equation}
\theta^*\in\arg\min_{\theta'\in\mathcal E_{\epsilon,\mathcal D}(\theta)} C_c(\theta').
\label{eq:app-generic}
\end{equation}
If $D_{\mathcal D}$ is separating over the intended function domain, $\epsilon=0$ recovers exact functional equivalence; under a finite sampled audit, zero empirical distortion remains sample-conditioned. Because $C_c$ is external and algorithmic, it is used only for exact evaluation and monotone accept/reject decisions.

\subsection{A two-by-two strict-gauge toy example}
\label{app:toy-gauge}
\begin{figure}[t]
\centering
\begin{tikzpicture}[
    font=\small,
    factor/.style={draw=black!35, rounded corners=2pt, fill=black!1,
                   inner sep=6pt, text width=4.15cm, minimum height=4.65cm,
                   align=center},
    summary/.style={draw=black!25, rounded corners=2pt, fill=black!2,
                    inner sep=6pt, text width=12.7cm, align=center},
    >=Latex]

\node[factor, anchor=north west] (orig) at (0,0) {
  {\bfseries Original factors}\\[0.45em]
  $A=\begin{bmatrix}2&4\\4&8\end{bmatrix}$\\[0.7em]
  $B=\begin{bmatrix}\tfrac12&\tfrac14\\1&\tfrac12\end{bmatrix}$\\[0.75em]
  \rule{3.45cm}{0.3pt}\\[-0.1em]
  \scriptsize Distinct values: $\{\tfrac14,\tfrac12,1,2,4,8\}$\\[0.2em]
  $H_{\rm symbol}=2.5$ bits/value
};

\node[factor, anchor=north east] (reg) at (13.55,0) {
  {\bfseries More repetitive factors}\\[0.45em]
  $A'=\begin{bmatrix}1&2\\1&2\end{bmatrix}$\\[0.7em]
  $B'=\begin{bmatrix}1&1\\2&2\end{bmatrix}$\\[0.75em]
  \rule{3.45cm}{0.3pt}\\[-0.1em]
  \scriptsize Distinct values: $\{1,2\}$\\[0.2em]
  $H_{\rm symbol}=1.0$ bits/value
};

\node[align=center, text width=4.05cm] (gauge) at (6.78,-1.15) {
  {\bfseries Exact power-of-two gauge}\\[0.45em]
  $M=\operatorname{diag}(2,4)$\\[0.35em]
  $A'=M^{-1}A,\qquad B'=BM$
};

\draw[->, line width=0.9pt] (4.50,-3.25) -- (9.05,-3.25);

\node[summary, anchor=north] (same) at (6.78,-5.05) {
  $BA=B'A'=\begin{bmatrix}2&4\\4&8\end{bmatrix}$\\[0.45em]
  {\bfseries Exactly the same linear map, with more repetitive stored values}\\[0.15em]
  \scriptsize Symbol entropy is illustrative only; the actual search minimizes measured compressed bytes.
};
\end{tikzpicture}
\caption{\textbf{Toy strict gauge.} A power-of-two diagonal gauge preserves the product while reducing six stored values to two; entropy is illustrative and selection uses real codec bytes.}
\label{fig:toy-gauge}
\end{figure}
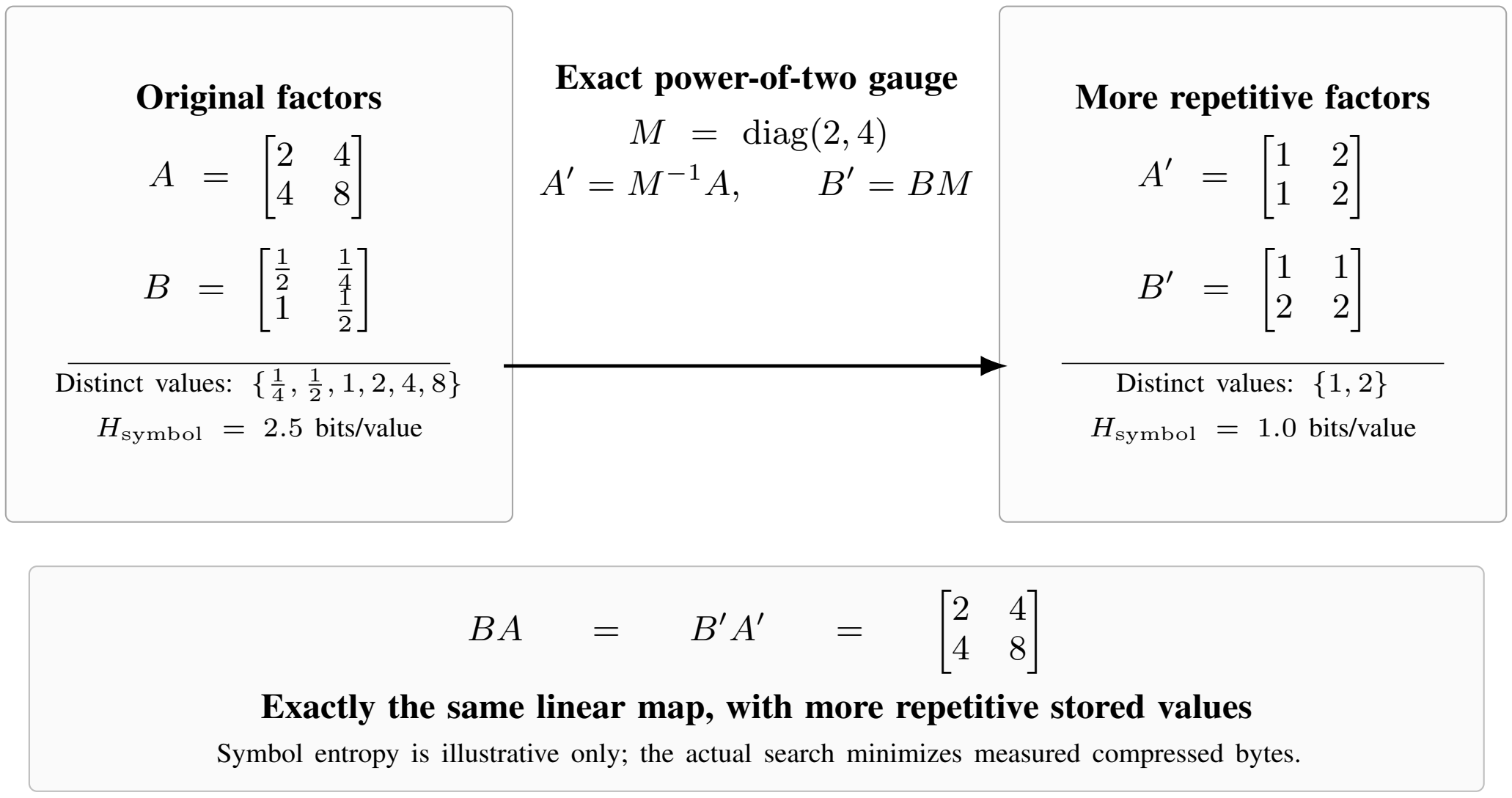

The simplest intuition is already visible in a two-layer linear map $y=BAx$. For any invertible hidden-space transform $M$, the low-rank factor reparameterization from Table~\ref{tab:exact-gauges},
\begin{equation}
A'=M^{-1}A,\qquad B'=BM
\end{equation}
preserves the composed operator exactly because $B'A'=BA$. Concretely, choose
\begin{align}
A&=\begin{bmatrix}2&4\\4&8\end{bmatrix}, &
B&=\begin{bmatrix}\tfrac12&\tfrac14\\1&\tfrac12\end{bmatrix}, &
M&=\operatorname{diag}(2,4),
\label{eq:toy-gauge-values}\\
A'=M^{-1}A&=\begin{bmatrix}1&2\\1&2\end{bmatrix}, &
B'=BM&=\begin{bmatrix}1&1\\2&2\end{bmatrix}.
\label{eq:toy-gauge-factors}
\end{align}
The product is unchanged explicitly:
\begin{equation}
BA=B'A'=\begin{bmatrix}2&4\\4&8\end{bmatrix}.
\label{eq:toy-gauge-product}
\end{equation}
Thus the eight entries of $(A,B)$ use six distinct values, $\{\tfrac14,\tfrac12,1,2,4,8\}$, whereas $(A',B')$ uses only $\{1,2\}$; their empirical symbol entropies are $2.5$ and $1.0$ bits/value. Figure~\ref{fig:toy-gauge} visualizes the calculation. Entropy is only a repetition proxy: strict-gauge search evaluates serialized candidates with the physical codec and accepts them only when bytes decrease and the declared exactness audit passes. The toy isolates the mechanism---same function, different representation, different low-level repetition---without conflating the method with an entropy model.

Because the entries of $M$ are powers of two, the scaling itself is exactly representable in binary floating-point. That fact does not by itself guarantee bitwise-identical execution after arbitrary finite-precision reductions, so the model-scale strict regime still uses the stored-precision audit defined below. The example is intentionally illustrative rather than a codec benchmark: with only eight scalar entries, container and codec headers would dominate any meaningful physical-byte comparison.

\subsection{SwiGLU permutation and reciprocal-scaling gauges}
Consider a bias-free SwiGLU MLP, where SiLU denotes the sigmoid linear unit,
\begin{align}
g &= W_g x, & u &= W_u x,\\
h &= \operatorname{SiLU}(g)\odot u, & y &= W_d h.
\end{align}
Let $P$ be a permutation matrix over the intermediate channels and $D$ an invertible diagonal matrix. A function-preserving transform is
\begin{equation}
W_g' = P W_g,\qquad
W_u' = D P W_u,\qquad
W_d' = W_d P^{-1}D^{-1}.
\label{eq:swiglu-gauge}
\end{equation}
Because SiLU and the elementwise product commute with permutation, and because $D$ acts on the linear branch after permutation,
\begin{align}
h' &= \operatorname{SiLU}(PW_gx)\odot DPW_ux\\
   &= DP\left(\operatorname{SiLU}(W_gx)\odot W_ux\right)=DPh,
\end{align}
so $W_d'h'=W_dh$. In strict stored-precision experiments we restrict $D$ to exactly representable power-of-two scalings and audit the realized BF16 computation after storage.

\subsection{Attention value/output basis gauge}
For a value subspace associated with a key/value head $h$, write the relevant value and output maps as $V_h$ and $O_{q\leftarrow h}$, where multiple query heads $q$ may share the same KV head. For any invertible basis transform $\Gamma_h$,
\begin{equation}
V_h' = \Gamma_h^{-1}V_h,
\qquad
O_{q\leftarrow h}'=O_{q\leftarrow h}\Gamma_h,
\label{eq:vo-gauge}
\end{equation}
which preserves the composed value/output operator because
\begin{equation}
O_{q\leftarrow h}'V_h'=O_{q\leftarrow h}\Gamma_h\Gamma_h^{-1}V_h
=O_{q\leftarrow h}V_h.
\end{equation}
Group permutations are applied consistently to queries, keys, values, and matching output-projection channels. Arbitrary dense $\Gamma_h$ is algebraically exact but can change stored-precision reductions after casting back to the stored dtype; such cases are separated from strict equality by an explicit finite-precision or data-conditioned audit.

\subsection{Low-rank factor gauge}
For a low-rank update $\Delta W=BA$, any invertible matrix $M$ in the latent rank space gives
\begin{equation}
B'=BM,\qquad A'=M^{-1}A,
\end{equation}
and therefore $B'A'=BA$. This gauge applies to static adapters and low-rank checkpoint histories; our experiments distinguish injected gauge drift from naturally continuous optimizer trajectories.

\subsection{Finite-precision audit tiers}
Table~\ref{tab:exactness-tiers-main} summarizes the three audit tiers in the main paper. Algebraic equality need not imply bitwise equality after BF16/FP16 storage and reordered reductions. In detail, we separate:
\begin{enumerate}
  \item \textbf{strict stored-precision audit}: sampled maximum absolute and relative output errors are both exactly zero;
  \item \textbf{finite-precision tolerance}: sampled stored-precision error is below a declared numerical threshold;
  \item \textbf{data-conditioned functional neighborhood}: activation/logit/NLL metrics are constrained on real text rather than on synthetic probes.
\end{enumerate}
This distinction avoids relabeling an algebraically exact transform as bitwise exact when finite-precision execution changes the realized computation.

\subsection{Activation-aware local distortion}
For a linear map $W\in\mathbb R^{m\times n}$ and calibration inputs $X\in\mathbb R^{n\times T}$, define the empirical Gram matrix
\begin{equation}
G=XX^\top/T.
\end{equation}
The local functional distortion is
\begin{align}
D_{\mathrm{local}}(W,W')
&=\frac{\|(W'-W)X\|_F^2}{\|WX\|_F^2}\\
&=\frac{\operatorname{tr}\left[(W'-W)G(W'-W)^\top\right]}
        {\operatorname{tr}\left[WGW^\top\right]}.
\label{eq:app-local}
\end{align}
For SwiGLU, gate and up projections use the MLP-input second-moment matrix. The down projection uses a separate second-moment matrix from the post-SiLU-product intermediate activation, so the distortion is measured in the actual input geometry of each map.

\subsection{Free-weight proposal families}
The first proposal family uses a diagonal approximation $G\approx\operatorname{diag}(g)$ and solves
\begin{equation}
\min_{W'}\sum_{ij} g_j(W'_{ij}-W_{ij})^2+\lambda |W'_{ij}|.
\end{equation}
The coordinate solution is weighted soft thresholding,
\begin{equation}
W_{ij}'=\operatorname{sign}(W_{ij})
\max\left(|W_{ij}|-\frac{\lambda}{2g_j},0\right),
\label{eq:soft-threshold}
\end{equation}
which creates exact zeros. The diagonal approximation is used only for proposal generation; admissibility is re-evaluated with the full empirical Gram matrix in Eq.~\ref{eq:app-local}.

The second family applies a significand-snapping operator $Q_b$ that rounds each value to a grid retaining $p-b$ significand bits while preserving tensor shape and dtype. The implementation counts $p=8,11,24,53$ significand bits for BF16, FP16, FP32, and FP64, respectively. Writing a finite scalar in \texttt{frexp} form as $w=m\,2^e$, the implementation computes
\begin{equation}
Q_b(w)=2^e\frac{\operatorname{round}(m2^{p-b})}{2^{p-b}},
\qquad b\ge 1.
\label{eq:significand-snap}
\end{equation}
Thus ``1-bit'' means $b=1$ in this snapping operator; it does \emph{not} mean that the tensor is stored as a 1-bit integer. The multibit setting generates separate $Q_b$ candidates for $b\in\{1,2,3,4\}$. This operation can increase repetition in low-level floating-point patterns but is not assumed to be function preserving. It must pass the selected functional trust region and exact physical reranking.

\paragraph{BF16 toy for $Q_1$ and $Q_2$.}
BF16 stores seven explicit fraction bits plus the implicit leading one, so the normalized significand has $p=8$ bits. Consider
\begin{equation}
\begin{aligned}
W&=\begin{bmatrix}1.2109375&1.234375\\1.2421875&1.2578125\end{bmatrix},\\
Q_1(W)&=\begin{bmatrix}1.21875&1.234375\\1.25&1.25\end{bmatrix},\qquad
Q_2(W)=\begin{bmatrix}1.21875&1.25\\1.25&1.25\end{bmatrix}.
\end{aligned}
\label{eq:bf16-snap-toy}
\end{equation}
All four entries lie in $[1,2)$ and therefore share sign bit $0$ and exponent field \texttt{01111111}. Their stored seven-bit fraction fields are
\begin{center}
\begin{tabular}{lcccc}
\toprule
& $W_{11}$ & $W_{12}$ & $W_{21}$ & $W_{22}$ \\
\midrule
$W$ & \texttt{0011011} & \texttt{0011110} & \texttt{0011111} & \texttt{0100001} \\
$Q_1(W)$ & \texttt{0011100} & \texttt{0011110} & \texttt{0100000} & \texttt{0100000} \\
$Q_2(W)$ & \texttt{0011100} & \texttt{0100000} & \texttt{0100000} & \texttt{0100000} \\
\bottomrule
\end{tabular}
\end{center}
Thus $Q_b$ rounds to a coarser significand grid rather than applying a literal bit mask: rounding may propagate a carry (for example, \texttt{0011111}$\to$\texttt{0100000}). Here $Q_1$ already creates repeated BF16 values and $Q_2$ creates more, exposing byte-level regularity without changing dtype.

For the activation-aware $\ell_1$ family, let $g_j=G_{jj}$ and define scores $a_{ij}=2|W_{ij}|g_j$. For target sparsity $s$, we take $k=\lceil s mn\rceil$ and choose $\lambda$ as the $k$th order statistic of the scores. Equation~\ref{eq:soft-threshold} then creates approximately the requested fraction of exact zeros. The token-free weight-only baseline uses the same construction with a uniform metric $g_j=1$.

Importantly, \textbf{the significand-snap and $\ell_1$ candidates form a union, not a composed transform}. For each tensor, the candidate generator evaluates
\begin{equation}
\mathcal P(W_0)=\{Q_b(W_0):b\in\mathcal B\}\cup\{L_s(W_0):s\in\mathcal S\},
\end{equation}
with all candidates constructed from the immutable original tensor $W_0$; the current tensor only removes duplicates and defines the physical baseline. Later sweeps may replace $Q_1(W_0)$ by $Q_2(W_0)$, or an $\ell_1$ candidate by a significand candidate, but never recursively round an already modified tensor. Dense RD experiments use $\mathcal S=\{5\times10^{-4},10^{-3},2\times10^{-3},5\times10^{-3}\}$. We call $\mathcal B=\{1\}$ \emph{1-bit+$\ell_1$} and $\mathcal B=\{1,2,3,4\}$ \emph{multibit+$\ell_1$}; legacy internal experiment logs may use the label \texttt{mantissa-L1}; the label denotes this union, not a composed operator.

\begin{table}[t]
\caption{Free-weight proposal semantics. ``Bit'' denotes the number of removed low-order significand bits; it is not a quantized deployment bitwidth.}
\label{tab:proposal-semantics}
\centering
\begin{tabular}{llll}
\toprule
Proposal & Construction & Shape/dtype & Exact? \\
\midrule
$Q_1$ & 1-bit significand snap & unchanged & no \\
$Q_{1:4}$ & alternatives $Q_1,\ldots,Q_4$ & unchanged & no \\
$L_s$ & $\ell_1$ soft threshold, target $s$ & unchanged & no \\
Exact gauge & algebraic reparameterization & unchanged & yes (algebraic) \\
\bottomrule
\end{tabular}
\end{table}

\subsection{Block and logit trust regions}
For a complete MLP block $F_{\mathrm{MLP}}$, we use a normalized block-output distortion
\begin{equation}
D_{\mathrm{block}}=
\frac{\|F_{\mathrm{MLP}}(X;\theta')-F_{\mathrm{MLP}}(X;\theta)\|_F}
     {\|F_{\mathrm{MLP}}(X;\theta)\|_F}.
\end{equation}
For final logits, let $p_i$ and $q_i$ be the original and candidate next-token distributions at position $i$. We report
\begin{equation}
D_{\mathrm{KL}}^{\mathrm{mean}}
=\frac{1}{N}\sum_{i=1}^N \mathrm{KL}(p_i\|q_i),
\end{equation}
together with high token-wise KL quantiles and causal negative-log-likelihood drift
\begin{equation}
\Delta\mathrm{NLL}=\mathrm{NLL}(\theta')-\mathrm{NLL}(\theta).
\end{equation}
Cumulative behavior metrics always compare with the immutable original model, preventing accepted small changes from resetting the trust region.

\subsection{Proposal--audit--rerank algorithm}
The implementation separates \emph{proposal construction}, \emph{physical scoring}, and \emph{behavioral auditing}. The dense free-weight path applies a cheap local filter, evaluates whole-model compressed bytes, then spends a forward pass only on physically improving candidates. Every commit must satisfy both the behavior constraint and strict physical improvement. Disjoint validation occurs only after endpoint freeze and never selects updates; its measured distortion defines the held-out RD coordinate rather than certifying exact transfer of the calibration limit.

Algorithm~\ref{lst:candidate-generation} summarizes dense candidate construction. Using $W_0$ keeps proposal depth interpretable across sweeps and prevents recursive numerical error.

\begin{lstlisting}[caption={Dense free-weight candidate construction. $W_0$ is the immutable original tensor; $W$ is the current committed tensor.},label={lst:candidate-generation}]
GenerateCandidates(W0, W, bit_set B, sparsities S, Gram G):
    P = empty set

    for s in S:
        cand = DiagonalL1(W0, G, target_sparsity=s)
        if cand differs from W0 and W:
            if LocalDistortion(W0, cand, G) <= eps_local:
                add cand to P

    for b in B:
        cand = SnapSignificand(W0, depth=b)   # Q_b(W0)
        if cand differs from W0 and W:
            if LocalDistortion(W0, cand, G) <= eps_local:
                add cand to P

    remove duplicate byte strings
    sort P by (local compressed bytes, local distortion)
    return first scan_limit candidates
\end{lstlisting}

Algorithm~\ref{lst:physical-search-main} in the main paper shows the coordinate search used for the output-functional experiments. The \texttt{exact-candidates-per-weight} parameter controls how many generated alternatives receive whole-model codec evaluation; setting it to zero means all candidates surviving the scan limit are evaluated. In the multibit+$\ell_1$ experiments this value is zero, so aggressive $Q_3/Q_4$ candidates cannot exclude safer $Q_1/Q_2$ alternatives through candidate truncation.

For exact-gauge search the outer physical-selection logic is the same, but candidate generation is algebraic and the behavior predicate is replaced by the declared exactness audit. If $\theta_t$ is the current state and $\widehat\theta$ is the selected admissible candidate, then
\begin{equation}
\theta_{t+1}=\begin{cases}
\widehat\theta, & C_c(\widehat\theta)<C_c(\theta_t),\\
\theta_t, & \text{otherwise.}
\end{cases}
\end{equation}
Thus $C_c$ is non-increasing without requiring derivatives.

\subsection{Formal Motivation and Guarantees}
\label{app:theory}

This section formalizes the representation-selection view used throughout the paper. The statements are intentionally modest: they explain why functional non-identifiability creates a compression domain, what the empirical rate--distortion curves mean, and what the greedy physical search guarantees. They do not claim global optimality for the implemented finite proposal families or generalization beyond the measured data.

\subsubsection{Compression as representative selection on a functional quotient}
Let $\Theta$ be the parameter space and define functional equivalence by
\begin{equation}
\theta\sim_f\theta' \quad\Longleftrightarrow\quad f_\theta=f_{\theta'}.
\end{equation}
The represented neural function is therefore associated with an equivalence class
$[\theta]_f=\{\vartheta\in\Theta:\vartheta\sim_f\theta\}$ rather than with a unique parameter vector. In contrast, the physical storage cost
\begin{equation}
C_c(\theta)=|c(S(\theta))|
\end{equation}
depends on the serialized parameter bytes. In general, $C_c$ is \emph{not} constant on $[\theta]_f$ and therefore does not descend to a well-defined scalar on the quotient space $\Theta/{\sim_f}$. This mismatch is the basic opportunity exploited by exact function-preserving compression.

Suppose a symmetry family $\mathcal G$ acts on parameters through $T_g:\Theta\rightarrow\Theta$ and satisfies $f_{T_g(\theta)}=f_\theta$ for every admissible $g\in\mathcal G$. Its orbit
\begin{equation}
\mathcal O_{\mathcal G}(\theta)=\{T_g(\theta):g\in\mathcal G\}
\subseteq [\theta]_f
\end{equation}
is the exact search domain available to that gauge family. The codec-specific orbit rate is
\begin{equation}
R_c^{\mathcal G}(\theta)=\inf_{g\in\mathcal G} C_c(T_g(\theta)).
\label{eq:orbit-rate}
\end{equation}
Different serializers or codecs can prefer different representatives because they induce different $C_c$ on the same orbit; no single gauge need be universally canonical for compression.

\paragraph{Proposition 1 (exact-orbit safety).}
If the identity transform belongs to $\mathcal G$, then $R_c^{\mathcal G}(\theta)\le C_c(\theta)$. Any selected $T_g(\theta)$ remains functionally equivalent to $\theta$ in exact arithmetic. Moreover, any finite composition of such exact transforms remains in $[\theta]_f$.

\paragraph{Proof.}
The identity is a feasible member of the infimum in Eq.~\ref{eq:orbit-rate}, giving the rate inequality. Functional equality follows from the defining symmetry property, and transitivity of equality gives the composition statement. \hfill$\square$

The proposition guarantees safety, not a strict byte improvement: if $C_c$ happens to be constant over the explored orbit, exact gauge search has zero compression headroom. The natural-LoRA negative results in Appendix~\ref{app:negative} are an empirical example of this boundary.

\subsubsection{Functional rate--distortion as an enlargement of the quotient class}
Let $D(f_{\theta'},f_{\theta_0})\ge 0$ be a functional distortion with $D(f_{\theta_0},f_{\theta_0})=0$. The ideal codec-specific rate--distortion envelope is
\begin{equation}
R_c^\star(\delta;\theta_0)
=
\inf_{\theta'\in\Theta:\,D(f_{\theta'},f_{\theta_0})\le\delta}
C_c(\theta').
\label{eq:ideal-rd}
\end{equation}
When $D$ is separating---that is, $D(f,g)=0$ only when $f=g$---Eq.~\ref{eq:ideal-rd} at $\delta=0$ reduces to optimization over exact functional equivalence; increasing $\delta$ enlarges the feasible set.

\paragraph{Proposition 2 (monotonicity of the ideal envelope).}
If $0\le\delta_1\le\delta_2$, then
\begin{equation}
R_c^\star(\delta_2;\theta_0)\le R_c^\star(\delta_1;\theta_0).
\end{equation}

\paragraph{Proof.}
The feasible set for $\delta_1$ is contained in the feasible set for $\delta_2$. Taking an infimum over a superset cannot increase the optimum. \hfill$\square$

The implemented method searches only a finite, structured subset $\mathcal H\subset\Theta$ generated by exact gauges, significand snaps, and $\ell_1$ proposals. Its best attainable rate
\begin{equation}
R_{c,\mathcal H}^\star(\delta;\theta_0)
=
\inf_{\theta'\in\mathcal H:\,D(f_{\theta'},f_{\theta_0})\le\delta} C_c(\theta')
\end{equation}
satisfies $R_{c,\mathcal H}^\star\ge R_c^\star$. Thus, for any fixed distortion definition and evaluation sample, an achieved endpoint is feasible at its own measured distortion and therefore gives an \emph{upper bound on the unknown minimum achievable byte cost} for that empirical problem, not a certificate of the globally optimal neural compression frontier.

The experiments additionally separate search-time and report-time distortions. For a calibration limit $\epsilon$, let
\begin{equation}
\widehat\theta_\epsilon
=
\mathcal A\!\left(\theta_0;\widehat D_{\mathrm{cal}}\le\epsilon,C_c\right)
\end{equation}
denote the frozen endpoint returned by the search algorithm $\mathcal A$. The plotted empirical RD point is
\begin{equation}
\left(
\widehat D_{\mathrm{val}}(f_{\widehat\theta_\epsilon},f_{\theta_0}),
C_c(\widehat\theta_\epsilon)
\right),
\label{eq:measured-rd-point}
\end{equation}
where validation is disjoint from candidate selection. Consequently, $\epsilon$ indexes a search run but is not itself the reported distortion coordinate. The finite greedy algorithm also need not produce points that are monotone in $\epsilon$ after held-out remeasurement; the non-dominated subset of the points defined by Eq.~\ref{eq:measured-rd-point} is the achieved empirical envelope.

\subsubsection{Monotone physical descent and finite termination}
For coordinate $k$, let $\mathcal P_k$ be the finite set of candidates generated from the immutable original tensor at that coordinate. At a current state $\theta$, define the one-coordinate neighborhood
\begin{equation}
\mathcal N(\theta)=
\left\{\theta^{(k\leftarrow p)}: k\in\{1,\ldots,K\},\ p\in\mathcal P_k\right\},
\end{equation}
and let $\mathcal N_\epsilon(\theta)\subseteq\mathcal N(\theta)$ contain the candidates that pass the declared cumulative behavior or exactness audit relative to the original model. The physical search commits only $\theta'\in\mathcal N_\epsilon(\theta)$ satisfying
\begin{equation}
C_c(\theta')<C_c(\theta).
\end{equation}

\paragraph{Proposition 3 (finite strict descent).}
Assume $C_c$ is measured as a nonnegative integer number of bytes and every accepted update strictly decreases it. If $C_{\min}$ is any lower bound on the attainable byte count, then the number of accepted updates is at most
\begin{equation}
C_c(\theta_0)-C_{\min}.
\end{equation}
Hence the accepted-update sequence cannot cycle and must terminate after finitely many commits.

\paragraph{Proof.}
A strict decrease between integer byte counts reduces $C_c$ by at least one byte per accepted update. A sequence of more than $C_c(\theta_0)-C_{\min}$ such decreases would fall below the lower bound. \hfill$\square$

Surrogate quality does not affect the guarantee: poor surrogates may waste evaluations or miss regions, but accepted steps cannot increase bytes because the real codec is evaluated before commitment.

\paragraph{Proposition 4 (coordinatewise local optimality of an exhaustive no-accept sweep).}
Suppose a sweep evaluates every member of $\mathcal N_\epsilon(\theta)$ and accepts none. Then
\begin{equation}
C_c(\theta)\le C_c(\vartheta)
\qquad
\text{for every }\vartheta\in\mathcal N_\epsilon(\theta),
\end{equation}
so $\theta$ is a one-coordinate local optimum with respect to the declared proposal family and audit.

\paragraph{Proof.}
If some feasible one-coordinate neighbor had smaller physical cost, the exhaustive sweep would evaluate it and the strict-improvement acceptance rule would permit a commit. \hfill$\square$

When a scan limit truncates the generated set, the same statement applies only to the scanned neighborhood. Likewise, coordinatewise local optimality does not imply a global optimum: two individually inadmissible or non-improving edits may become useful when applied jointly. The LoRA-history experiments in Appendix~\ref{app:negative} motivate this type of interaction.

\subsubsection{Why the activation Gram matrix is the local functional geometry}
The activation-aware metric in Eq.~\ref{eq:app-local} follows directly from the expected perturbation of a linear map. Let $\Delta W=W'-W$ and let $x$ be a random layer input with second moment $\Sigma=\mathbb E[xx^\top]$. Then
\begin{align}
\mathbb E\|\Delta W x\|_2^2
&=\mathbb E\!\left[x^\top\Delta W^\top\Delta W x\right]\\
&=\operatorname{tr}\!\left(\Delta W\,\Sigma\,\Delta W^\top\right).
\label{eq:expected-linear-distortion}
\end{align}
For calibration activations $X=[x_1,\ldots,x_T]$, the empirical second moment $G=XX^\top/T$ is therefore the plug-in estimator of this exact linear-layer output distortion. Normalizing by $\operatorname{tr}(WGW^\top)$ yields the dimensionless relative criterion used in Eq.~\ref{eq:app-local}.

Approximating $G$ by $\operatorname{diag}(g)$ makes the quadratic proposal objective separable across scalar weights. Adding an $\ell_1$ penalty gives
\begin{equation}
\min_{w'} g(w'-w)^2+\lambda|w'|,
\end{equation}
whose minimizer is the weighted soft-threshold operator in Eq.~\ref{eq:soft-threshold}. Thus activation-aware $\ell_1$ is the closed-form proximal solution under a diagonal approximation to the measured local functional geometry, not an arbitrary sparsifier. The approximation only proposes candidates; block/logit audits and physical codec evaluation remain authoritative.

\subsubsection{Why mean token KL is a useful output-space distortion}
For each token position $i$, let $p_i$ and $q_i$ be the original and candidate next-token distributions. With natural logarithms, Pinsker's inequality gives
\begin{equation}
\operatorname{TV}(p_i,q_i)
\le
\sqrt{\tfrac{1}{2}\operatorname{KL}(p_i\|q_i)},
\end{equation}
where $\operatorname{TV}(p,q)=\tfrac12\|p-q\|_1$. Averaging over $N$ positions and applying Jensen's inequality yields
\begin{equation}
\frac1N\sum_{i=1}^N\operatorname{TV}(p_i,q_i)
\le
\sqrt{\frac{1}{2N}\sum_{i=1}^N\operatorname{KL}(p_i\|q_i)}
=
\sqrt{\tfrac12 D_{\mathrm{KL}}^{\mathrm{mean}}}.
\label{eq:mean-kl-tv}
\end{equation}
Thus mean token KL controls average one-step distributional deviation but not argmax identity, long-horizon rollouts, or downstream accuracy; small shifts can cross decision boundaries and compound after generation leaves the original prefix. We therefore report top-1 agreement, NLL/perplexity, tail KL, cross-domain C4, and downstream benchmarks separately.

\subsubsection{Scope of the guarantees}
The theory above separates three kinds of statements. Exact algebraic gauges preserve the represented function by construction; strict stored-precision claims remain tied to the declared execution audit. The physical search guarantees monotone byte descent and, under exhaustive sweeps, a coordinatewise local optimum over the implemented proposal neighborhood. Functional RD points are empirical operating points measured on held-out data, not guarantees for arbitrary unseen sequences and not proofs of the global envelope in Eq.~\ref{eq:ideal-rd}. This separation is deliberate: the mathematical structure motivates and constrains the search, while the real codec and independent behavioral measurements determine the reported empirical outcome.

\section{Experimental Details}
\label{app:experiments}

\subsection{Models and serialization}
The principal model is SmolLM2-135M \citep{allal2025smollm2}; SmolLM2-360M provides scale validation, and Qwen3-0.6B-Base \citep{yang2025qwen3} provides architecture transfer from a distinct grouped-query-attention (GQA) family. Every physical comparison uses deterministic sorted-key serialization, and whole-model optimization recompresses the full stream after each candidate patch. Module-framed experiments are reported only to study local versus global physical objectives.

\subsection{Codec specification and provenance}
The primary codec is Zstandard level 9; gzip and Brotli provide cross-codec evaluation. Backend and version are part of each experimental condition: the same 269,030,016-byte SmolLM2-135M tensor stream compresses to 208,502,655 bytes under Zstandard CLI v1.5.5 but 209,332,798 bytes under v1.4.8 on another system. Absolute byte counts are therefore never mixed across these environments; cross-method gains use a pinned codec build.

\subsection{Calibration, validation, and test splits}
Table~\ref{tab:data-use-protocol} summarizes the split roles. Activation-aware search uses WikiText-2-raw-v1 with 32 sequences of length 256 for calibration (8,192 tokens) and an independent 8,192-token validation sample. Saved free-weight endpoints are evaluated post hoc on 8,192 tokens from the test split. For cross-domain auditing, C4 \texttt{en} validation is streamed only until an 8,192-token sample is assembled; that sample is cached, and C4 is never used for endpoint search. For SmolLM2-135M, this corresponds to approximately 5.33 calibration token samples per dimension for the 1,536-dimensional down-projection input and 14.22 for the 576-dimensional MLP input. For SmolLM2-360M, the corresponding ratios are 3.20 for the 2,560-dimensional down-projection input and 8.53 for the 960-dimensional MLP input.

\begin{table}[!h]
\caption{Data-use protocol for the LLM experiments. Each listed stage uses 32 sequences of length 256 (8,192 tokens) when run.}
\label{tab:data-use-protocol}
\centering
\setlength{\tabcolsep}{4pt}
\begin{tabular}{@{}lll@{}}
\toprule
Data source & Role & Selects updates? \\
\midrule
WikiText-2 raw train & calibration / behavior gate & yes \\
WikiText-2 raw validation & held-out RD measurement & no \\
WikiText-2 raw test & final in-domain audit & no \\
C4 \texttt{en} validation & streamed cross-domain audit & no \\
\bottomrule
\end{tabular}
\end{table}

\subsection{Representation classes}
\paragraph{Exact gauge.}
We optimize SwiGLU permutations/reciprocal power-of-two scalings and grouped-query-attention permutations/value-output gauges. Strict experiments require zero sampled stored-precision output error.

\paragraph{Finite-precision gauge.}
The same algebraic families are admitted under a declared BF16 numerical tolerance rather than strict sampled equality.

\paragraph{Weight-space free weights.}
The proposal families are identical to the activation-aware methods, but admissibility uses normalized parameter MSE only and requires no calibration data.

\paragraph{Local activation.}
Candidates must satisfy the full-Gram local reconstruction criterion in Eq.~\ref{eq:app-local}.

\paragraph{Block activation.}
Candidates satisfy local reconstruction and end-of-MLP output constraints.

\paragraph{Output-space functional rate--distortion.}
Each distortion point restarts from the immutable original model. The primary distortion coordinate is mean next-token KL between the original and candidate output distributions; high token-wise KL quantiles, NLL/perplexity, and top-1 agreement are reported separately. Measured validation KL defines the held-out RD coordinate.

Qwen3 architecture transfer follows the same split and physical-objective semantics, with six independent search limits ($5\times10^{-4}$ through $8\times10^{-3}$) and the same 1-bit+$\ell_1$ proposal family. Its measured validation KL, not the requested calibration limit, is used in Figure~\ref{fig:llm-rd-scale}.

Experiments~106 and~104 use matched 135M/360M multibit+$\ell_1$ pure-KL semantics: calibration mean KL alone controls acceptance, with max-token-KL and absolute-NLL gates set to $10^9$; validation remains post-search and never selects updates. The historical guarded 135M 1-bit+$\ell_1$ and multibit+$\ell_1$ endpoints are retained separately for provenance and the earlier whole-file serialization audit. Experiments~107 and~105 provide matched 135M and 360M cross-codec transfer, respectively; Experiments~108 and~109 provide the corresponding post hoc behavioral audits. Experiment~109 evaluates all thirteen 360M endpoints on 8,192 WikiText test tokens, 8,192 streamed C4 tokens, and five zero-shot downstream tasks. Each downstream baseline/endpoint pair runs sequentially within one NVIDIA A40 allocation, eliminating heterogeneous-GPU variation from the paired comparison.

The earlier local+logit sweep additionally applied a full-Gram local activation threshold before evaluating KL/NLL; it is therefore reported separately from the local-gate-off output-functional frontier. In these dense runs, ``1-bit+$\ell_1$'' means significand-snap depth set $\{1\}$ plus the default $\ell_1$ sparsity proposals, while ``multibit+$\ell_1$'' expands the significand-snap depth set to $\{1,2,3,4\}$; neither term denotes integer quantization.

\subsection{Cross-codec protocol}
We report two cross-codec protocols. The legacy 135M representation-class matrix optimizes each class independently for Zstandard, gzip, and Brotli, then cross-evaluates every endpoint under all three. The main functional-RD transfer lets only Zstandard select frozen 135M/360M pure-KL endpoints; gzip, Brotli, and tensor-aware ZipNN then remeasure those states without refinement. Older scale controls with short target-codec refinement keep all behavior constraints referenced to the immutable original model, not an intermediate Zstandard endpoint.

\section{Detailed Experimental Evidence}
\label{app:results}

\subsection{Exact whole-model gauges}
Independent module-framed optimization initially produced a 67,971-byte reduction on SmolLM2-135M, but only 6,194 bytes survived when the entire model was serialized and compressed as one stream. Moving the whole-model codec into the coordinate-descent loop increased the converged strict-equality gain to 14,416 bytes. A finite-precision tolerance tier reached 16,154 bytes. The mismatch between module-framed and monolithic optima demonstrates that a physical objective cannot in general be decomposed into a sum of local codec objectives.

\subsection{Activation-aware free-weight hierarchy}
Under Zstandard CLI v1.5.5, the completed controlled study gives the following gains for Zstandard-optimized endpoints:
\begin{table}[ht]
\caption{Hierarchy of Zstandard-optimized representation classes on SmolLM2-135M. Cross-codec values are obtained without reoptimizing the endpoint.}
\centering
\begin{tabular}{lrrr}
\toprule
Method & Zstd & gzip & Brotli \\
\midrule
Strict gauge & 14,416 & 6,134 & 5,039 \\
Finite-precision gauge & 16,154 & 13,503 & 8,776 \\
Weight-space & 32,471 & 24,984 & 36,688 \\
Block activation & 253,728 & 225,237 & 271,357 \\
Local activation & 1,432,379 & 1,237,083 & 1,480,472 \\
\bottomrule
\end{tabular}
\end{table}
The local activation criterion exposes about 99 times the Zstandard gain of the strict exact-gauge endpoint, while the block-constrained criterion exposes about 18 times the exact-gauge gain.

\subsection{Held-out endpoint behavior}
A separate Zstandard v1.4.8 run saved complete free-weight endpoints and evaluated them on 8,192 untouched WikiText test tokens. Because the codec build differs, the byte gains from this run are not numerically merged with the v1.5.5 table above; the behavioral metrics are reported as an independent replication of the same method ordering.

\begin{table}[ht]
\caption{Held-out behavior of saved SmolLM2-135M free-weight endpoints. $\Delta$PPL is relative to baseline perplexity 29.3441.}
\centering
\begin{tabular}{lrrrrr}
\toprule
Endpoint & $\Delta$NLL & $\Delta$PPL (\%) & mean KL & P99.9 KL & top-1 agreement \\
\midrule
Weight-space & $+3.33\!\times\!10^{-4}$ & +0.0333 & 0.001456 & 0.0516 & 97.28\% \\
Local activation & $+6.13\!\times\!10^{-4}$ & +0.0613 & 0.001669 & 0.0481 & 97.50\% \\
Block activation & $+2.22\!\times\!10^{-4}$ & +0.0222 & 0.001286 & 0.0426 & 97.71\% \\
\bottomrule
\end{tabular}
\end{table}
The block-activation endpoint improves physical gain while lowering every reported held-out distortion statistic relative to weight-space. The local endpoint is more aggressive---it changes more tensors and incurs more mean distortion---but keeps the perplexity increase below 0.1\%.

\begin{figure}[ht]
  \centering
  \includegraphics[width=0.68\linewidth]{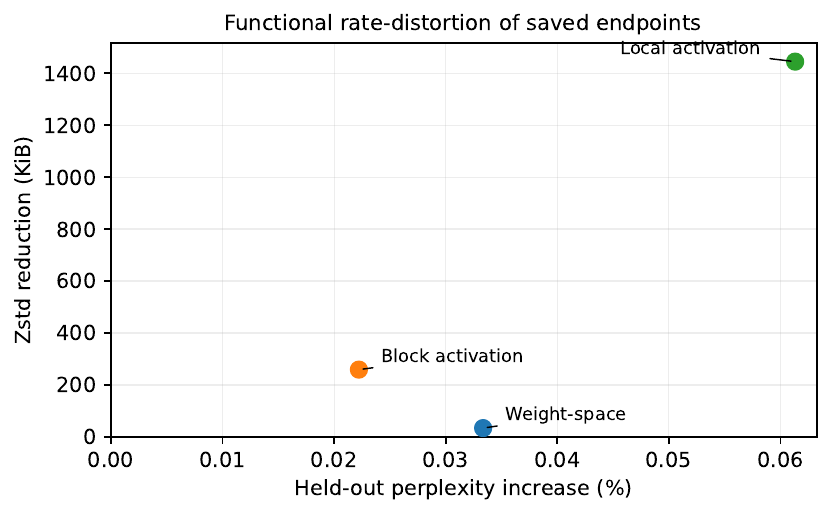}
  \caption{Zstandard gain versus held-out perplexity change for saved free-weight endpoints from one pinned run. Values come directly from recorded endpoint files.}
  \label{fig:functional-tradeoff}
\end{figure}

\subsection{Exhaustive method--codec matrix}
Table~\ref{tab:codec-matrix-full} reports the complete 21-run study. Each row is optimized for one codec and cross-evaluated under the other two. Some gzip/Brotli runs hit the sweep cap before an explicit zero-gain sweep; these rows are marked, and byte differences of only hundreds or thousands should not be over-interpreted.

\begin{table}[ht]
\caption{Complete SmolLM2-135M method--codec matrix. Entries are byte gains under the evaluation codec; bold marks the optimization codec.}
\label{tab:codec-matrix-full}
\centering
\begin{tabular}{llrrr}
\toprule
Method & Optimized for & Zstd gain & gzip gain & Brotli gain \\
\midrule
Block activation & Zstd & \textbf{253,728} & 225,237 & 271,357 \\
Block activation & gzip & 253,330 & \textbf{225,361} & 271,015 \\
Block activation & Brotli & 253,227 & 225,043 & \textbf{270,984} \\
\midrule
Strict gauge & Zstd & \textbf{14,416} & 6,134 & 5,039 \\
Strict gauge & gzip & 3,344 & \textbf{14,487} & 6,570 \\
Strict gauge & Brotli & 2,482 & 10,501 & \textbf{15,801} \\
\midrule
Finite-precision gauge & Zstd & \textbf{16,154} & 13,503 & 8,776 \\
Finite-precision gauge & gzip & 5,882 & \textbf{19,678} & 10,877 \\
Finite-precision gauge & Brotli & 2,613 & 13,182 & \textbf{22,061} \\
\midrule
Local activation & Zstd & \textbf{1,432,379} & 1,237,083 & 1,480,472 \\
Local activation & gzip & 1,430,444 & \textbf{1,235,934} & 1,479,248 \\
Local activation & Brotli & 1,431,300 & 1,235,829 & \textbf{1,481,369} \\
\midrule
Weight-space & Zstd & \textbf{32,471} & 24,984 & 36,688 \\
Weight-space & gzip & 32,341 & \textbf{25,677} & 36,511 \\
Weight-space & Brotli & 32,345 & 24,410 & \textbf{37,391} \\
\midrule
Logit (initial strict gate) & Zstd & \textbf{0} & 0 & 0 \\
Logit (initial strict gate) & gzip & 0 & \textbf{0} & 0 \\
Logit (initial strict gate) & Brotli & 0 & 0 & \textbf{0} \\
Joint (initial strict gate) & Zstd & \textbf{0} & 0 & 0 \\
Joint (initial strict gate) & gzip & 0 & \textbf{0} & 0 \\
Joint (initial strict gate) & Brotli & 0 & 0 & \textbf{0} \\
\bottomrule
\end{tabular}
\end{table}

\begin{figure}[ht]
  \centering
  \includegraphics[width=0.82\linewidth]{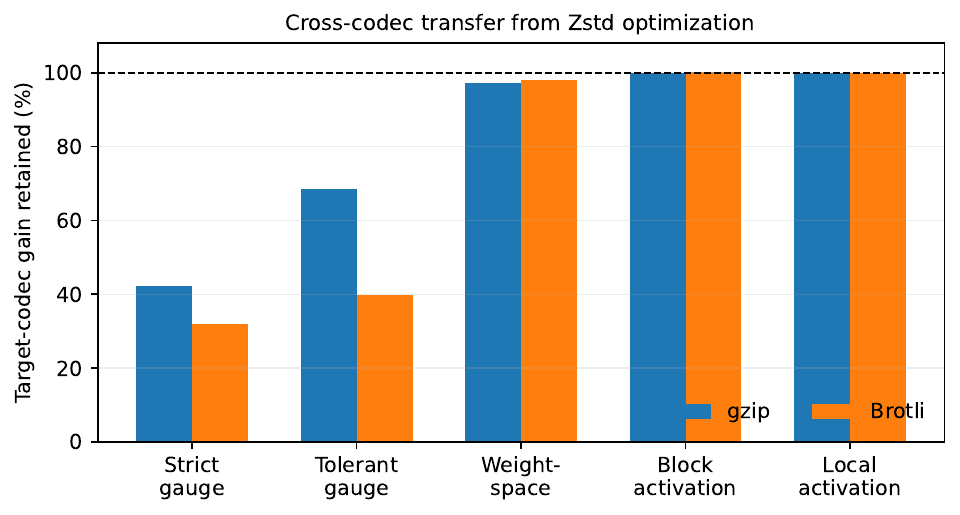}
  \caption{Target-codec gain from Zstandard-selected endpoints. Activation-aware endpoints transfer strongly; exact gauges remain codec-specific; values above 100\% reflect path-dependent target-codec search.}
  \label{fig:codec-transfer}
\end{figure}

\subsection{ZipNN specialized-codec transfer}
Experiment~95 evaluates saved Zstandard-selected endpoints under ZipNN 0.5.4 with its Huffman safetensors path. ZipNN is not used to select or refine any endpoint. Every state is first canonicalized to the same sorted-key single-file safetensors representation, every size-reducing compressed tensor is decompressed and checked bit-for-bit against an immutable pre-compression snapshot, and gains are measured relative to the untouched model compressed by the identical ZipNN configuration. The real-backend BF16 preflight passes at both scales. ZipNN 0.5.4 mutates its disposable compression inputs, but immutable-reference verification remains unaffected: all 272 floating tensors at 135M and all 290 at 360M are successfully compressed and verified for every row with zero fallbacks.

\begin{table}[ht]
\caption{Post hoc ZipNN 0.5.4 transfer of Zstandard-selected SmolLM2 endpoints; positive gain indicates transfer.}
\label{tab:zipnn-transfer}
\centering
\begin{tabular}{lrrrr}
\toprule
Representation & 135M ZipNN bytes & 135M gain & 360M ZipNN bytes & 360M gain \\
\midrule
Original & 179,380,579 & 0 & 481,538,381 & 0 \\
Strict gauge & 179,380,323 & +256 & 481,530,639 & +7,742 \\
Weight-space & 179,385,542 & -4,963 & 481,563,747 & -25,366 \\
Block activation & 179,119,181 & +261,398 & 481,175,152 & +363,229 \\
Local activation & 177,943,210 & +1,437,369 & 474,275,441 & +7,262,940 \\
\bottomrule
\end{tabular}
\end{table}

The scale extension reproduces the pattern. Block-activation ZipNN gain rises from 261,398 bytes at 135M to 363,229 at 360M; local gain grows from 1,437,369 to 7,262,940 bytes. At 360M these retain about 91.5\% and 90.1\% of the corresponding Zstandard gains (397,107 and 8,064,543 bytes), although ZipNN never ranks candidates. Weight-space perturbations instead enlarge the ZipNN representation at both scales, while exact gauges remain much weaker than activation-aware shaping. Thus ZipNN supports codec complementarity and transferable representation shaping, not replacement or superiority over the codec.

\subsection{Whole-file safetensors serialization check}
\label{app:whole-safetensors}
The primary generic-codec experiments use deterministic tensor serialization inside the evaluation pipeline. To verify that the measured gain survives a standard deployable checkpoint container, we additionally materialize the untouched SmolLM2-135M model and the two saved output-functional endpoints as complete single-file \texttt{.safetensors} artifacts and compress each file as one opaque byte stream. Table~\ref{tab:whole-safetensors} reports the resulting file sizes. The raw serialized checkpoints are essentially unchanged in nominal size: the optimized files are only 32 bytes smaller than the 269,060,552-byte original. The substantial reductions therefore appear only after lossless compression and cannot be explained by a smaller tensor inventory, shape change, or lower-precision deployment format.

\begin{table*}[b]
\caption{Whole-file SmolLM2-135M serialization. Zstandard, gzip, and Brotli compress each complete \texttt{.safetensors} file as one opaque stream; percentages are reductions from the original.}
\label{tab:whole-safetensors}
\centering
\footnotesize
\setlength{\tabcolsep}{1.5pt}
\begin{tabular}{lrrrr}
\toprule
Checkpoint & Raw safetensors & Zstd file & gzip file & Brotli file \\
\midrule
Original & 269,060,552 & 208,496,205 & 213,332,680 & 209,496,002 \\
1-bit+$\ell_1$ & 269,060,520 & 195,630,069 (6.17\%) & 203,141,714 (4.78\%) & 197,136,946 (5.90\%) \\
guarded multibit+$\ell_1$ & 269,060,520 & 165,110,519 (20.81\%) & 169,332,666 (20.63\%) & 165,302,676 (21.10\%) \\
\bottomrule
\end{tabular}
\end{table*}

This whole-file check complements controlled codec accounting: gains remain after ordinary serialization introduces headers, metadata, and cross-tensor byte-stream interactions. Because the listing records artifact sizes but not generic-codec executable/library versions or hashes, we treat it as deployment-style validation rather than a version-normalized codec benchmark; Experiment~103 retains the raw and normalized records.

\clearpage
\subsection{SmolLM2-360M scale validation}
Experiment~94 repeats the four most informative methods on SmolLM2-360M with the same 8,192-token calibration and validation protocol, then audits saved endpoints on 8,192 untouched WikiText test tokens. The run uses Zstandard CLI v1.5.5, matching the 135M matrix. Table~\ref{tab:llm-scale-main} gives the cross-model Zstandard/ZipNN summary and held-out 360M behavior.

The 360M Zstandard baseline is 560,070,491 bytes versus 208,502,655 at 135M (2.686$\times$). For strict-gauge, weight-space, block-activation, and local-activation endpoints, gzip gains are 12,004/49,039/316,621/6,309,191 bytes and Brotli gains 4,729/73,136/392,380/7,901,788 bytes. Strict-gauge gain is nearly unchanged (14,205 versus 14,416 bytes); weight, block, and local gains grow 2.77$\times$, 1.57$\times$, and 5.63$\times$, with local reaching 8,064,543 bytes (1.4399\%). Its perplexity ratio is 1.00081285 (+0.0813\%), mean KL 0.00151764, P99.9 KL 0.0318769, and top-1 agreement 97.7173\%. Two sizes provide scale validation, not a scaling law.

\subsection{Output-space rate--distortion and matched controls}
Experiment~93 isolates an important distinction between a hard local-activation proposal filter and a pure output-space trust region. The initial \texttt{MODE=logit} sweep retained a local activation threshold of $10^{-5}$ before evaluating KL/NLL, so it is properly a \emph{local+logit-constrained} frontier rather than a pure output-space frontier. A subsequent local-gate-off 1-bit+$\ell_1$ run kept the same whole-model Zstandard objective, two sweeps, and $|\Delta\mathrm{NLL}|\le2\times10^{-3}$, but made the local proposal/validation limits effectively inactive. With a permissive mean-KL limit of 0.1, all 90 eligible MLP projection tensors accepted one edit in sweep~1 and sweep~2 found no further byte-improving move. The Zstandard v1.5.5 compressed size falls from the 208,502,655-byte baseline to 195,660,703 bytes, a gain of 12,841,952 bytes (6.1591\%). Despite the permissive limit, the endpoint naturally lands at validation mean KL 0.00185345 and validation $\Delta$NLL $+2.14\times10^{-4}$.

The held-out WikiText/C4 audit is summarized in Table~\ref{tab:llm-heldout-main} in the main body. Post hoc ZipNN 0.5.4 in Huffman mode compresses the untouched canonical state to 179,380,579 bytes and the same Zstandard-selected endpoint to 167,612,739 bytes, a gain of 11,767,840 bytes (6.5603\% of the ZipNN-compressed original). This retains 91.64\% of the Zstandard byte gain. All 272 floating tensors are successfully compressed and verified bit-for-bit after round trip, with zero fallbacks.

The older local+logit frontier remains useful for showing how a stricter local notion of admissibility contracts the physical search space. Table~\ref{tab:logit-rd-validation} records that completed sweep. The $2\times10^{-3}$ limit accepts 12 physical improvements and saves 1,434,249 bytes, ending at validation mean KL $1.6759\times10^{-3}$ and validation $\Delta$NLL $+6.80\times10^{-4}$. Tightening the limit to $10^{-3}$ leaves three accepted updates and 258,412 bytes of gain at mean KL $9.4690\times10^{-4}$. At $6\times10^{-4}$, the run accepts no update from the original model. Since each threshold restarts from the original state and the lower thresholds only tighten the same deterministic mean-KL gate, the $3\times10^{-4}$ and $10^{-4}$ controls cannot admit an update once the $6\times10^{-4}$ control accepts none; they are therefore omitted rather than interpolated.

\begin{table}[ht]
\caption{SmolLM2-135M local+logit validation sweep with a $10^{-5}$ local-activation prefilter. KL and NLL are measured on the optimizer validation split.}
\label{tab:logit-rd-validation}
\centering
\begin{tabular}{rrrrrr}
\toprule
Mean-KL limit & Zstd gain & accepted & val. mean KL & val. P99.9 KL & val. $\Delta$NLL \\
\midrule
$2.0\times10^{-3}$ & 1,434,249 & 12 & 0.00167586 & 0.0731186 & +0.000679794 \\
$1.0\times10^{-3}$ & 258,412 & 3 & 0.000946901 & 0.00779131 & +0.000242316 \\
$6.0\times10^{-4}$ & 0 & 0 & 0 & 0 & 0 \\
\bottomrule
\end{tabular}
\end{table}

\begin{figure}[ht]
  \centering
  \includegraphics[width=0.68\linewidth]{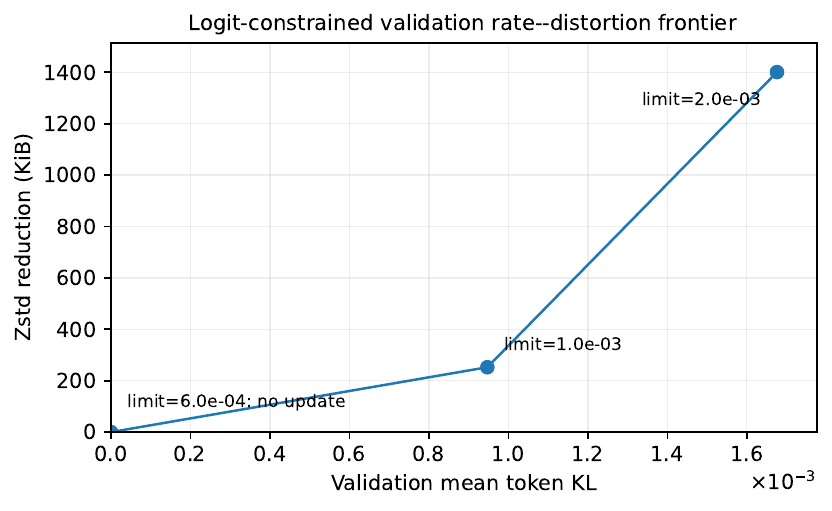}
  \caption{Local+logit validation frontier with a $10^{-5}$ local-activation proposal filter. The local-gate-off 1-bit+$\ell_1$ point isolates the filter's effect.}
  \label{fig:logit-rd-validation}
\end{figure}

The completed local-gate-off 1-bit+$\ell_1$ 135M frontier under Zstandard v1.5.7 is strongly discrete: measured gains are 12,851,750 B at the $2\times10^{-3}$ limit, 10,652,457 B at $1.8\times10^{-3}$, 8,152,470 B at $1.6\times10^{-3}$, 3,354,034 B at $1.4\times10^{-3}$, 2,623,192 B at $1.2\times10^{-3}$, 507,272 B at $10^{-3}$, 278,015 B at $8\times10^{-4}$, and zero at $6\times10^{-4}$. Because calibration and validation are disjoint, figures use the measured validation KL on the horizontal axis for every endpoint. The nominal calibration limit is retained only as search provenance; whether the held-out estimate lies numerically above or below that limit is not treated as RD pass/fail.

A matched SmolLM2-360M 1-bit+$\ell_1$ sweep under Zstandard v1.5.7 uses a 560,070,082-byte baseline. At a $2\times10^{-3}$ calibration-KL limit, the endpoint saves 39,839,878 B (7.1134\%), accepts 96 updates, and has validation mean KL 0.001805929 with $\Delta$NLL $-2.58\times10^{-5}$. A permissive $10^{-2}$ KL-only probe edits all 96 searchable tensors in sweep~1, reaches exactly the same gain, then completes a zero-accept sweep, identifying 39.84 MB as the current 1-bit+$\ell_1$ proposal-family ceiling. We do not report the full WikiText test/C4 suite for this endpoint; it establishes the one-bit ceiling and representative downstream behavior.

\paragraph{SmolLM2-135M multibit+$\ell_1$ pure-KL frontier.}
Experiment~106 records the corrected parallel 135M sweep after an earlier model-identity mismatch was resolved and the true SmolLM2-135M checkpoint was verified (hidden size 576, intermediate size 1,536, 30 layers; Zstandard v1.5.7 baseline 208,502,655~B). Each KL control is an independent restart from the original model. Candidate acceptance uses calibration mean token KL only; max-token-KL and absolute-NLL gates are set to $10^9$, and validation is post-search measurement only. Table~\ref{tab:multibit-135m-rd} reports all fourteen controls.

The corrected frontier first exceeds the 12.852-MB 1-bit+$\ell_1$ ceiling at the $3\times10^{-3}$ control, saving 14.836~MB (7.115\%) at held-out KL 0.003536. Its widest boundary saves 43.413~MB (20.821\%) at KL 0.010321 with validation $\Delta$NLL $+0.01203$, so we treat it as high-distortion rather than preferred. Experiment~106 supplies the 135M curve in Figure~\ref{fig:llm-rd-scale}; the earlier auxiliary-NLL sweep remains separate because its downstream and whole-file audits use historical checkpoints.

\begin{table}[ht]
\caption{Complete corrected SmolLM2-135M multibit+$\ell_1$ pure-KL sweep. Each row is an independent restart.}
\label{tab:multibit-135m-rd}
\centering
\begin{tabular}{rrrrrr}
\toprule
Search KL & Zstd gain & Reduction (\%) & Accepted & Val. mean KL & Val. $\Delta$NLL \\
\midrule
$2.0\times10^{-4}$ & 0 B & 0.000\% & 0 & 0 & 0 \\
$4.0\times10^{-4}$ & 0 B & 0.000\% & 0 & 0 & 0 \\
$6.0\times10^{-4}$ & 0.128 MB & 0.062\% & 1 & 0.0005136 & -0.0004320 \\
$8.0\times10^{-4}$ & 0.129 MB & 0.062\% & 1 & 0.0007464 & -0.0009368 \\
$1.0\times10^{-3}$ & 0.874 MB & 0.419\% & 5 & 0.0009862 & -0.0019206 \\
$1.5\times10^{-3}$ & 4.643 MB & 2.227\% & 25 & 0.0015587 & -0.0005922 \\
$2.0\times10^{-3}$ & 6.694 MB & 3.211\% & 25 & 0.0020983 & +0.0008426 \\
$2.5\times10^{-3}$ & 9.783 MB & 4.692\% & 30 & 0.0028557 & +0.0015110 \\
$3.0\times10^{-3}$ & 14.836 MB & 7.115\% & 45 & 0.0035355 & +0.0026570 \\
$4.0\times10^{-3}$ & 25.888 MB & 12.416\% & 71 & 0.0043785 & +0.0034335 \\
$5.0\times10^{-3}$ & 33.301 MB & 15.972\% & 84 & 0.0053522 & +0.0036164 \\
$6.0\times10^{-3}$ & 38.958 MB & 18.685\% & 88 & 0.0064986 & +0.0073230 \\
$8.0\times10^{-3}$ & 43.149 MB & 20.695\% & 90 & 0.0084364 & +0.0075504 \\
$1.0\times10^{-2}$ & 43.413 MB & 20.821\% & 90 & 0.0103206 & +0.0120284 \\
\bottomrule
\end{tabular}
\end{table}

\paragraph{Matched 135M pure-KL cross-codec transfer.}
Experiment~107 evaluates every Experiment~106 endpoint post hoc under gzip, Brotli, and tensor-aware ZipNN while Zstandard alone selects the states. At the $3\times10^{-3}$ control, Zstd/gzip/Brotli/ZipNN reductions are 7.12/6.88/7.23/7.66\%; at $5\times10^{-3}$ they are 15.97/15.65/16.25/16.92\%; and at the $10^{-2}$ boundary they are 20.82/20.68/21.24/21.95\%. ZipNN bitwise-verifies all 272 floating tensors with zero fallbacks at every endpoint.

\begin{table}[t]
\caption{Post hoc transfer across the SmolLM2-135M matched pure-KL multibit+$\ell_1$ frontier. Endpoints are Zstandard-selected; gzip, Brotli, and ZipNN never enter search.}
\label{tab:multibit-135m-cross-codec}
\centering
\setlength{\tabcolsep}{3.2pt}
\begin{tabular}{rrrrrr}
\toprule
Search KL & Val. mean KL & Zstd & gzip & Brotli & ZipNN \\
\midrule
0.0002 & 0 & 0.000\% & 0.000\% & 0.000\% & 0.000\% \\
0.0004 & 0 & 0.000\% & 0.000\% & 0.000\% & 0.000\% \\
0.0006 & 0.000514 & 0.062\% & 0.053\% & 0.064\% & 0.073\% \\
0.0008 & 0.000746 & 0.062\% & 0.053\% & 0.064\% & 0.073\% \\
0.0010 & 0.000986 & 0.419\% & 0.362\% & 0.425\% & 0.470\% \\
0.0015 & 0.001559 & 2.227\% & 1.959\% & 2.246\% & 2.456\% \\
0.0020 & 0.002098 & 3.211\% & 3.005\% & 3.253\% & 3.501\% \\
0.0025 & 0.002856 & 4.692\% & 4.519\% & 4.760\% & 5.046\% \\
0.0030 & 0.003536 & 7.115\% & 6.880\% & 7.233\% & 7.661\% \\
0.0040 & 0.004379 & 12.416\% & 12.063\% & 12.614\% & 13.218\% \\
0.0050 & 0.005352 & 15.972\% & 15.651\% & 16.252\% & 16.918\% \\
0.0060 & 0.006499 & 18.685\% & 18.426\% & 19.031\% & 19.720\% \\
0.0080 & 0.008436 & 20.695\% & 20.552\% & 21.101\% & 21.832\% \\
0.0100 & 0.010321 & 20.821\% & 20.683\% & 21.236\% & 21.953\% \\
\bottomrule
\end{tabular}
\end{table}

\paragraph{SmolLM2-360M multibit+$\ell_1$ pure-KL frontier.}
Experiment~104 records the completed parallel multibit+$\ell_1$ sweep under the same Zstandard v1.5.7 baseline. Each KL control is an independent restart from the original model. Candidate acceptance uses calibration mean token KL only; the max-token-KL and absolute-NLL gates are set to $10^9$, so these diagnostics do not constrain search. Validation remains post-search and never selects updates. Table~\ref{tab:multibit-360m-rd} reports all thirteen controls using measured held-out validation mean KL as the distortion coordinate.

The positive 360M multibit+$\ell_1$ points form a monotone physical frontier as measured held-out KL increases. The larger proposal family still does not dominate the recorded 1-bit+$\ell_1$ greedy search at tight distortion: for example, the multibit+$\ell_1$ $2\times10^{-3}$ control saves 22.37~MB at validation KL 0.001971, whereas the 1-bit+$\ell_1$ endpoint saves 39.84~MB at KL 0.001806. This is consistent with path dependence of the greedy coordinate search and does not imply a worse attainable optimum for the superset proposal family. The first multibit+$\ell_1$ point above the 1-bit+$\ell_1$ proposal-family ceiling occurs at the $3\times10^{-3}$ control, saving 65.65~MB (11.72\%) at held-out KL 0.002904. The widest tested boundary saves 131.09~MB (23.41\%) at held-out KL 0.008871 with validation $\Delta$NLL $+0.00783$; we therefore treat it as a high-distortion RD boundary rather than a preferred preservation point. Experiment~105 records post hoc transfer for every endpoint under gzip, Brotli, and tensor-aware ZipNN, while Zstandard alone selects the states. At the $3\times10^{-3}$ control the Zstandard, gzip, Brotli, and ZipNN reductions are 11.72/11.08/11.83/12.24\%; at $5\times10^{-3}$ they are 20.56/20.02/20.91/21.44\%; and at the $10^{-2}$ boundary they are 23.41/23.09/23.89/24.43\%. ZipNN compresses all 290 floating tensors; bitwise round-trip verification succeeds with zero fallbacks at every endpoint. Experiment~109 additionally audits all thirteen endpoints on WikiText test, streamed C4, and A40-matched downstream tasks; the three zero-edit controls reproduce the baseline exactly, providing an end-to-end audit sanity check.

\paragraph{Qwen3-0.6B architecture transfer.}
Experiment~99 applies the identical 1-bit+$\ell_1$ dense search to \texttt{Qwen/Qwen3-0.6B-Base} under Zstandard v1.5.7. From a 922,297,440-B compressed baseline, the $5\times10^{-4}$ and $10^{-3}$ searches save 514,280 B (0.0558\%) and 2,088,469 B (0.2264\%) at validation mean KL 0.000417 and 0.000934. At $2\times10^{-3}$, all 84 gate/up/down MLP projections across 28 layers accept the implementation-labeled \texttt{mantissa\_clear\_1} candidate, saving 44,288,844 B (4.8020\%) at validation mean KL 0.00163106 and $\Delta$NLL $+0.00108194$; the second sweep accepts nothing. Independent $3\times10^{-3}$, $5\times10^{-3}$, and $8\times10^{-3}$ restarts select the identical ordered 84-edit sequence and byte trajectory. All six controls are retained in CSV/JSON evidence; the repeated plateau identifies exhaustion of the current proposal family, not an intrinsic RD optimum.

\begin{table}[ht]
\caption{Complete SmolLM2-360M multibit+$\ell_1$ pure-KL sweep. Each row is an independent restart; zero-edit controls are retained explicitly.}
\label{tab:multibit-360m-rd}
\centering
\begin{tabular}{rrrrrr}
\toprule
Search KL & Zstd gain & Reduction (\%) & Accepted & Val. mean KL & Val. $\Delta$NLL \\
\midrule
$2.0\times10^{-4}$ & 0 B & 0.000\% & 0 & 0 & 0 \\
$4.0\times10^{-4}$ & 0 B & 0.000\% & 0 & 0 & 0 \\
$6.0\times10^{-4}$ & 0 B & 0.000\% & 0 & 0 & 0 \\
$8.0\times10^{-4}$ & 0.403 MB & 0.072\% & 1 & 0.0006521 & +0.0003250 \\
$1.0\times10^{-3}$ & 1.595 MB & 0.285\% & 4 & 0.0009426 & +0.0004730 \\
$1.5\times10^{-3}$ & 13.662 MB & 2.439\% & 25 & 0.0014127 & +0.0003701 \\
$2.0\times10^{-3}$ & 22.368 MB & 3.994\% & 21 & 0.0019706 & +0.0001586 \\
$3.0\times10^{-3}$ & 65.651 MB & 11.722\% & 67 & 0.0029043 & +0.0009295 \\
$4.0\times10^{-3}$ & 76.294 MB & 13.622\% & 73 & 0.0038812 & +0.0018085 \\
$5.0\times10^{-3}$ & 115.123 MB & 20.555\% & 98 & 0.0048806 & +0.0012246 \\
$6.0\times10^{-3}$ & 127.767 MB & 22.813\% & 97 & 0.0057489 & +0.0026548 \\
$8.0\times10^{-3}$ & 130.745 MB & 23.344\% & 97 & 0.0074679 & +0.0046200 \\
$1.0\times10^{-2}$ & 131.085 MB & 23.405\% & 96 & 0.0088707 & +0.0078297 \\
\bottomrule
\end{tabular}
\end{table}

\begin{table}[ht]
\caption{Post hoc transfer across the SmolLM2-360M multibit+$\ell_1$ frontier. Endpoints are Zstandard-selected; gzip, Brotli, and ZipNN never enter search.}
\label{tab:multibit-360m-cross-codec}
\centering
\begin{tabular}{rrrrrr}
\toprule
Search KL & Val. mean KL & Zstd & gzip & Brotli & ZipNN \\
\midrule
$2.0\times10^{-4}$ & 0 & 0.000\% & 0.000\% & 0.000\% & 0.000\% \\
$4.0\times10^{-4}$ & 0 & 0.000\% & 0.000\% & 0.000\% & 0.000\% \\
$6.0\times10^{-4}$ & 0 & 0.000\% & 0.000\% & 0.000\% & 0.000\% \\
$8.0\times10^{-4}$ & 0.0006521 & 0.072\% & 0.055\% & 0.071\% & 0.075\% \\
$1.0\times10^{-3}$ & 0.0009426 & 0.285\% & 0.219\% & 0.278\% & 0.302\% \\
$1.5\times10^{-3}$ & 0.0014127 & 2.439\% & 2.049\% & 2.404\% & 2.548\% \\
$2.0\times10^{-3}$ & 0.0019706 & 3.994\% & 3.805\% & 4.026\% & 4.170\% \\
$3.0\times10^{-3}$ & 0.0029043 & 11.722\% & 11.077\% & 11.830\% & 12.236\% \\
$4.0\times10^{-3}$ & 0.0038812 & 13.622\% & 13.046\% & 13.788\% & 14.223\% \\
$5.0\times10^{-3}$ & 0.0048806 & 20.555\% & 20.017\% & 20.907\% & 21.444\% \\
$6.0\times10^{-3}$ & 0.0057489 & 22.813\% & 22.418\% & 23.264\% & 23.801\% \\
$8.0\times10^{-3}$ & 0.0074679 & 23.344\% & 23.026\% & 23.828\% & 24.365\% \\
$1.0\times10^{-2}$ & 0.0088707 & 23.405\% & 23.093\% & 23.895\% & 24.428\% \\
\bottomrule
\end{tabular}
\end{table}

\begin{table}[ht]
\caption{Complete SmolLM2-360M pure-KL post hoc behavioral audit (Experiment~109). \emph{Max raw acc.} is the largest absolute endpoint-minus-original accuracy change across LAMBADA, HellaSwag, PIQA, ARC-Easy, and ARC-Challenge, in percentage points.}
\label{tab:multibit-360m-behavior}
\centering
\setlength{\tabcolsep}{3pt}
\begin{tabular}{@{}rrrrrr@{}}
\toprule
Search KL & Zstd & Wiki PPL $\Delta$ & C4 PPL $\Delta$ & Max raw acc. & LAMBADA PPL $\Delta$ \\
\midrule
$2.0\times10^{-4}$ & 0.000\% & +0.000\% & +0.000\% & 0.000 & +0.000\% \\
$4.0\times10^{-4}$ & 0.000\% & +0.000\% & +0.000\% & 0.000 & +0.000\% \\
$6.0\times10^{-4}$ & 0.000\% & +0.000\% & +0.000\% & 0.000 & +0.000\% \\
$8.0\times10^{-4}$ & 0.072\% & +0.028\% & +0.021\% & 0.380 & +0.022\% \\
$1.0\times10^{-3}$ & 0.285\% & -0.016\% & -0.059\% & 0.130 & -0.050\% \\
$1.5\times10^{-3}$ & 2.439\% & -0.041\% & +0.075\% & 0.260 & +0.047\% \\
$2.0\times10^{-3}$ & 3.994\% & +0.040\% & -0.195\% & 0.300 & -0.335\% \\
$3.0\times10^{-3}$ & 11.722\% & +0.219\% & +0.077\% & 0.430 & -0.273\% \\
$4.0\times10^{-3}$ & 13.622\% & +0.415\% & +0.150\% & 0.650 & +2.188\% \\
$5.0\times10^{-3}$ & 20.555\% & +0.307\% & +0.446\% & 0.760 & +3.594\% \\
$6.0\times10^{-3}$ & 22.813\% & +0.450\% & +0.242\% & 0.860 & +3.935\% \\
$8.0\times10^{-3}$ & 23.344\% & +0.668\% & +0.293\% & 0.490 & +2.351\% \\
$1.0\times10^{-2}$ & 23.405\% & +0.877\% & +0.570\% & 0.740 & +5.015\% \\
\bottomrule
\end{tabular}
\end{table}

Experiment~109 shows that the high-compression frontier remains close to the original model on the measured independent audits through the near-saturation region. At the $8\times10^{-3}$ control, Zstandard reduction is 23.344\% at held-out validation KL 0.007468; WikiText test and C4 perplexity increase by 0.668\% and 0.293\%, WikiText/C4 top-1 agreement is 95.19\%/95.41\%, the largest raw five-task accuracy deviation is 0.49 percentage points, and LAMBADA perplexity increases by 2.35\%. The $10^{-2}$ boundary reaches 23.405\% but adds only 340,224~B (0.061 percentage points) beyond $8\times10^{-3}$ while the LAMBADA perplexity increase grows from 2.35\% to 5.02\%. The full frontier's largest raw task-accuracy deviation is 0.86 percentage points (at the $6\times10^{-3}$ control), so these results are interpreted as preservation diagnostics rather than capability gains.

The matched 135M and 360M multibit+$\ell_1$ pure-KL sweeps appear in Figure~\ref{fig:llm-rd-scale}; Experiments~106 and~104 provide their RD evidence, Experiments~107 and~105 provide full 135M/360M cross-codec transfer, Experiment~108 adds the complete five-task 135M downstream audit, and Experiment~109 adds complete 360M WikiText test, C4, and five-task audits. The older guarded 135M frontier remains separate historical provenance for the earlier whole-file serialization study.

\subsection{Complete post hoc downstream benchmark audit}
\label{app:downstream}
\begin{figure*}[t]
  \centering
  \includegraphics[width=0.99\textwidth]{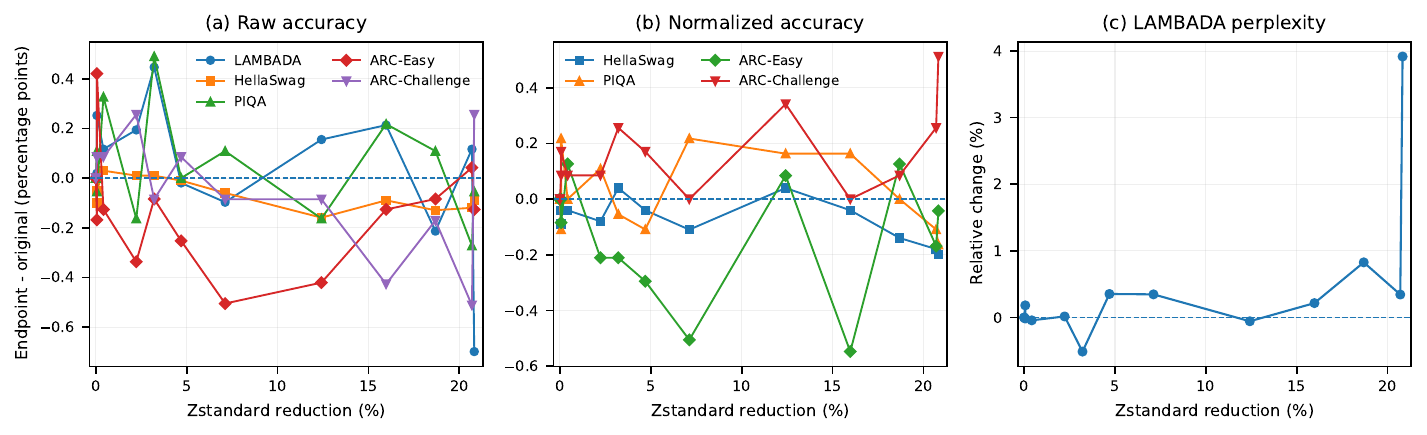}
  \caption{\textbf{Post hoc downstream preservation across the corrected SmolLM2-135M pure-KL multibit+$\ell_1$ frontier.} (a) Raw zero-shot accuracy deltas. (b) Normalized-accuracy deltas for four multiple-choice tasks. (c) Relative LAMBADA perplexity change. All fourteen independently restarted endpoints are shown versus measured Zstandard reduction; the sharp final-point perplexity rise occurs after the frontier has nearly saturated in physical byte savings.}
  \label{fig:multibit-downstream}
\end{figure*}

Experiment~100 audits representative frozen SmolLM2-135M/360M endpoints, Experiment~108 audits all 14 corrected 135M pure-KL multibit+$\ell_1$ endpoints, and Experiment~109 audits all 13 matched 360M endpoints. Experiments~100 and~109 are separate provenance records; absolute harness values are not mixed across them, and the 360M frontier deltas are taken only from within-run A40-matched comparisons. The full-frontier audits use zero-shot BF16 EleutherAI \texttt{lm-eval} with auto-selected batch size 64; for Experiment~109, every baseline/endpoint pair runs sequentially on the same NVIDIA A40. These tasks never influence compression search or selection. Tables~\ref{tab:downstream-135m-full}--\ref{tab:downstream-360m-full} give the earlier representative suites, Figure~\ref{fig:multibit-downstream} shows the full 135M frontier, and Table~\ref{tab:multibit-360m-behavior} summarizes the complete 360M frontier. Endpoint-minus-original deltas are paired by example; reported standard errors are aggregate uncertainties, not paired-difference confidence intervals.

\begin{table}[ht]
\caption{Complete SmolLM2-135M downstream audit. Values are mean $\pm$ harness-reported standard error.}
\label{tab:downstream-135m-full}
\centering
\setlength{\tabcolsep}{3pt}
\begin{tabular}{@{}llccc@{}}
\toprule
Task & Metric & Original & Endpoint & $\Delta$ \\
\midrule
ARC-Challenge & acc & $0.2824\pm0.0132$ & $0.2824\pm0.0132$ & +0.0000 \\
ARC-Challenge & acc\_norm & $0.2961\pm0.0133$ & $0.2969\pm0.0134$ & +0.0008 \\
ARC-Easy & acc & $0.6427\pm0.0098$ & $0.6410\pm0.0098$ & -0.0017 \\
ARC-Easy & acc\_norm & $0.5863\pm0.0101$ & $0.5875\pm0.0101$ & +0.0012 \\
HellaSwag & acc & $0.3559\pm0.0048$ & $0.3541\pm0.0048$ & -0.0018 \\
HellaSwag & acc\_norm & $0.4312\pm0.0049$ & $0.4301\pm0.0049$ & -0.0011 \\
LAMBADA & acc & $0.4287\pm0.0069$ & $0.4306\pm0.0069$ & +0.0019 \\
LAMBADA & perplexity & $19.0560\pm0.6292$ & $19.0044\pm0.6279$ & -0.0516 \\
PIQA & acc & $0.6812\pm0.0109$ & $0.6866\pm0.0108$ & +0.0054 \\
PIQA & acc\_norm & $0.6839\pm0.0108$ & $0.6839\pm0.0108$ & +0.0000 \\
\bottomrule
\end{tabular}
\end{table}

\begin{table}[ht]
\caption{Complete SmolLM2-360M downstream audit. Values are mean $\pm$ harness-reported standard error.}
\label{tab:downstream-360m-full}
\centering
\setlength{\tabcolsep}{3pt}
\begin{tabular}{@{}llccc@{}}
\toprule
Task & Metric & Original & Endpoint & $\Delta$ \\
\midrule
ARC-Challenge & acc & $0.3635\pm0.0141$ & $0.3660\pm0.0141$ & +0.0025 \\
ARC-Challenge & acc\_norm & $0.3831\pm0.0142$ & $0.3788\pm0.0142$ & -0.0043 \\
ARC-Easy & acc & $0.7033\pm0.0094$ & $0.7058\pm0.0094$ & +0.0025 \\
ARC-Easy & acc\_norm & $0.6814\pm0.0096$ & $0.6831\pm0.0095$ & +0.0017 \\
HellaSwag & acc & $0.4317\pm0.0049$ & $0.4304\pm0.0049$ & -0.0013 \\
HellaSwag & acc\_norm & $0.5637\pm0.0049$ & $0.5635\pm0.0049$ & -0.0002 \\
LAMBADA & acc & $0.5333\pm0.0070$ & $0.5337\pm0.0070$ & +0.0004 \\
LAMBADA & perplexity & $9.3938\pm0.2654$ & $9.3794\pm0.2648$ & -0.0144 \\
PIQA & acc & $0.7165\pm0.0105$ & $0.7187\pm0.0105$ & +0.0022 \\
PIQA & acc\_norm & $0.7149\pm0.0105$ & $0.7198\pm0.0105$ & +0.0049 \\
\bottomrule
\end{tabular}
\end{table}

The representative suites show small mixed-sign paired deltas. At 135M the largest raw-accuracy change is 0.54 percentage points (PIQA) and LAMBADA PPL decreases by 0.27\%; at 360M all raw-accuracy changes are within 0.25 points and LAMBADA PPL decreases by 0.15\%. These are small relative to the aggregate standard errors.

\paragraph{Downstream behavior across the corrected 135M pure-KL multibit+$\ell_1$ frontier.}
Experiment~108 evaluates all 14 Experiment~106 endpoints; materialization hashes match the same tensors recorded by Experiment~107. Through the $8\times10^{-3}$ control, Zstandard reduction is 20.695\% at held-out KL 0.008436, every raw-accuracy deviation is within 0.512 percentage points, and LAMBADA perplexity increases by only 0.345\%. The $10^{-2}$ boundary reaches 20.821\% at KL 0.010321 but decreases LAMBADA accuracy by 0.699 percentage points and increases perplexity by 3.918\%. Thus the last step buys only 263,680~B (0.126 percentage points) while worsening an independent language-modeling diagnostic, supporting its interpretation as an RD boundary rather than a preferred operating point. Mixed-sign task changes are not evidence of capability improvement.

\paragraph{Downstream behavior across the 360M pure-KL multibit+$\ell_1$ frontier.}
Experiment~109 evaluates all 13 Experiment~104 endpoints with hardware-matched baseline/endpoint pairs on NVIDIA A40 GPUs. The three zero-edit controls reproduce every downstream metric exactly at logged precision. At $8\times10^{-3}$, the largest raw-accuracy deviation is 0.49 percentage points and LAMBADA perplexity increases by 2.35\%; the $10^{-2}$ boundary decreases LAMBADA accuracy by 0.74 percentage points and increases perplexity by 5.02\% for only 0.061 additional percentage points of Zstandard reduction. Across the entire 13-point frontier, the largest raw-accuracy deviation is 0.86 percentage points. Together with the independent WikiText test and C4 audits in Table~\ref{tab:multibit-360m-behavior}, this completes the 360M behavioral audit while preserving the interpretation of the widest point as an RD boundary.

\section{Controlled Serializer and Systems Evidence}
\label{app:controlled}

The main paper emphasizes deployment-relevant whole-model transformer checkpoints. Controlled studies also test whether differentiable or structured functional geometry can improve discrete physical objectives across serializers and storage systems; they establish mechanism validity rather than headline LLM evidence.

\subsection{DEFLATE and gzip mechanism validation}
Experiment~02 constructs a DEFLATE-aware soft Bellman parser over RFC~1951 literal/length and distance actions while retaining the unmodified zlib/gzip implementation as the external evaluator. On the canonical seed-7 fixture, the soft fixed-tree objective correlates with the exact fixed-Huffman dynamic program at Pearson 0.99999986; the entropy-style readout correlates with real raw-DEFLATE size at Pearson 0.93587 and Spearman 0.83601, with 0.87097 pairwise ranking accuracy. A one-byte differentiable repair rounds back to the exact repeated byte and reduces real raw-DEFLATE size from 208 to 168 bits and deterministic gzip size from 352 to 312 bits. Across seeds 1--5, the reduced sweep reports mean raw-DEFLATE Pearson 0.99848, Spearman 0.98561, pairwise ranking 1.0, and successful exact-byte/gzip improvement on all five seeds.

Experiment~03 then combines the same rate model with an exact hidden-channel permutation symmetry in a two-layer MLP. A reversible channel-major layout serializes each hidden unit as one coupled record, and Sinkhorn optimization acts only on channel order. The six-channel fixture exhaustively evaluates all $6!=720$ hard permutations with the real gzip implementation, providing a controlled bridge from differentiable rate modeling to semantics-preserving representation selection.

\subsection{Delta coding and semantic checkpoint alignment}
Experiment~04 replaces within-object LZ matching with an immutable external base object and introduces a differentiable COPY/ADD parser. Experiment~05 combines that objective with exact hidden-channel permutation symmetry across base/target checkpoints. The synthetic target includes sparse high-magnitude decoys that deliberately mislead Euclidean alignment while changing only a few serialized bytes, so the benchmark tests compression-aware alignment rather than allowing $\ell_2$ matching to solve the task automatically. Final candidates are evaluated using an exact decodable DiffDelta patch, real Git binary-patch output, and optional VCDIFF when available.

\subsection{Content-defined chunking and deduplication}
Experiments~06--08 study a different non-differentiable storage mechanism: content-defined chunking and cross-version deduplication. The fully coupled SoftDedup study differentiates through both soft Rabin boundary probabilities and soft historical chunk reuse, but exact storage is always measured with unmodified Rabin boundaries and exact byte identity. On seed~7, the soft lifetime-storage objective reaches Pearson 0.999510 and Spearman 0.976190 against exact storage. Joint optimization restores both a lost chunk boundary and corrupted content, reducing exact physical storage from 66.25 to 33.75 bytes. On a semantics-preserving checkpoint trajectory, exact lifetime storage falls from 4,084 to 3,662 bytes with maximum function error $9.54\times10^{-7}$; the optimized trajectory matches the compression oracle. Across seeds 1--5, exact trajectory storage improves in all five runs, with a mean saving of 339.6 bytes over the unaligned trajectory.

\subsection{Structured Zstandard and Brotli surrogates}
Experiment~17 uses a Zstandard-specific sequence/rate surrogate with literal/match parsing, literal-length (LL), match-length (ML), and offset (OF) code classes, extra-bit accounting, recent offsets, and differentiable entropy readouts. On the seed-7 fixture it reaches Pearson 0.98916 and Spearman 1.0 against exact Zstandard bits; one-step repair reduces real output from 344 to 216 bits and recovers the repeated byte. Exact-byte repair and real-Zstandard improvement succeed for all seeds 1--5. Experiment~20 analogously uses a Brotli surrogate with soft LZ parsing, literal contexts, and selected static-dictionary opportunities while production Brotli remains the final evaluator.

Table~\ref{tab:controlled-systems-main} in the main paper summarizes representative measured outcomes across these mechanism studies. The detailed subsections here retain the protocol-specific context and caveats.

\subsection{Module-wise exact symmetry and serializer layout}
Experiment~85 evaluates the same mechanism on public Qwen3 modules and shows why exact physical reranking remains necessary. On a contiguous 32-channel SwiGLU window with reversible channel-major serialization, every one of eight random shared permutations improves exact Zstandard by 18--92 bytes while preserving function and leaving record-wise delta/dedup unchanged. Eight DiffZstd proposals also reduce exact Zstandard size, but the surrogate/exact Spearman correlation is $-0.193$ and the best differentiable-proposal gain (78 bytes) is smaller than the best random shared-permutation gain under the same evaluation budget (92 bytes). The lesson is not that the surrogate is universally reliable; rather, function-preserving representation choice measurably changes physical compression, while the production backend must remain authoritative.

These controlled studies motivate the architecture used in the main experiments: propose using differentiable or algebraic structure, audit functional admissibility, and rerank by the exact physical system. They also broaden the empirical scope beyond Zstandard/gzip/Brotli whole-model compression to delta coding, content-defined deduplication, and multiple codec-specific structured rate models.

\section{Negative Results and Limits of Functional Freedom}
\label{app:negative}

Functional non-identifiability creates an optimization domain, but it does not guarantee a useful compression gain. We retain several negative or weak results because they clarify where the proposed viewpoint is informative and where natural training already appears close to a stable representation.

\subsection{Natural module trajectories preserve channel identity}
In the public Qwen3-0.6B-Base $\rightarrow$ GSM8K-LoRA materialized trajectory, an eight-channel layer-14 delta probe gives 30,720 bytes for identity, target $\ell_2$ alignment, target SoftDelta alignment, and every shared permutation. Random target-only gauges are much worse (61,435--79,864 bytes). A 32-channel reciprocal power-of-two SwiGLU scale study similarly leaves identity best at 30,720 bytes, while even a one-channel $2^{-1}$ target-only change increases the objective to 34,819 bytes. These results indicate that the natural trajectory strongly preserves the dense model's latent-channel identity; exact relative-gauge alignment has little room to improve the delta representation.

\subsection{Static LoRA exact gauges provide only small codec gains}
A public rank-8 Qwen3 LoRA adapter exposes the exact factor gauge
\begin{equation}
BA=(BM)(M^{-1}A),\qquad M\in GL(r).
\end{equation}
In the public adapter probe, the untouched adapter is 4,287,504 bytes under Zstandard. A simple shared rank-balance gauge saves 838 bytes; the best local general-$GL(8)$ candidate saves 845 bytes, the best local QD candidate saves 857 bytes (0.01999\%), and a pure orthogonal candidate saves only 64 bytes. Thus exact rank-space freedom exists, but the static physical gain is small and dominated by factor balance rather than arbitrary dense rotation.

A synthetic positive control verifies that this is not a limitation of the gauge machinery. A deliberately dense non-monomial $GL(4)$ scramble leaves the LoRA operator unchanged but increases history cost. Signed-monomial, orthogonal, and QD alignment all remain at 6,162 bytes, whereas analytic full-$GL$ alignment reduces the objective to 2,744 bytes, removing 3,418 bytes (55.5\%) of artificial factor-basis cost. The public result is therefore best interpreted as a property of the natural trajectory rather than a failure to represent useful gauge transforms.

\subsection{Multi-checkpoint LoRA history is essentially gauge-stable}
The most diagnostic negative result is the five-checkpoint public Qwentaur trajectory
\[
\text{checkpoint-500}\rightarrow\text{1000}\rightarrow\text{1500}\rightarrow\text{1878}\rightarrow\text{final}.
\]
It contains 196 common rank-8 LoRA pairs per checkpoint. Anchoring the first snapshot leaves $4\times196=784$ mutable checkpoint/module gauge coordinates. The completed global-sparse exact search screens all 784 coordinates and then reranks the most promising proposals exactly against the authoritative full-history DiffDelta objective. After 2,057.36 seconds, 29 exact moves are accepted, reducing the objective from 60,567,188 to 60,567,091 bytes: only 97 bytes, or 0.000160\%.

This result is useful for two reasons. First, it prevents an overly broad claim that neural histories contain large arbitrary basis drift. Continuously trained LoRA factors can remain naturally gauge-stable. Second, it exposes a structural limitation of checkpoint-local history proposals: changing an intermediate snapshot affects both adjacent chronological deltas, so an improvement on one edge can cancel on the other. Joint temporal gauges may address this in future work, but the current paper does not claim a positive natural-history result.

\subsection{Why negative cases belong in the study}
The successful whole-model free-weight experiments and the weak natural LoRA-history results delimit two different sources of compressibility. The former exploit a broad neighborhood of numerically different representations that preserve behavior on data; the latter ask whether training happened to introduce arbitrary exact factor-basis drift. The evidence suggests that the first source can be substantial, while the second can be nearly absent. Reporting both cases makes the scope of ``compression over functional equivalence classes'' more precise: useful headroom depends on the geometry of the admissible class and on how the trained representation occupies that class.

\clearpage
\section{Reference Guide: Acronyms, Datasets, Models, and Software}
\label{app:reference-guide}

This appendix collects terminology and external artifacts used in the study so that the main text can remain compact. License labels below are informational summaries of the upstream repository or model- and dataset-card metadata checked on September 24, 2026; the upstream license text and terms of use remain authoritative. Synthetic fixtures created within this project are not external datasets and are therefore omitted from the dataset list.

\subsection{Acronyms, abbreviations, and compact notation}
\begingroup
\small
\setlength{\parskip}{0pt}

\paragraph{Acronyms, method names, and backend shorthand.}
\begin{description}
\setlength{\itemsep}{0.15em}
\item[AI] artificial intelligence.
\item[ARC] Allen Institute for AI (AI2) Reasoning Challenge; the paper uses the Easy and Challenge subsets.
\item[ASVD] activation-aware singular value decomposition.
\item[AWQ] activation-aware weight quantization.
\item[BF16] bfloat16 floating-point format.
\item[C4] Colossal Clean Crawled Corpus.
\item[CLI] command-line interface.
\item[CSV] comma-separated values; used for normalized tabular evidence records.
\item[DEFLATE] lossless LZ/Huffman compression format standardized by RFC~1951.
\item[DiffDelta] the project's exact delta-compression objective/backend.
\item[FP16] IEEE 754 binary16 half-precision floating-point format.
\item[FP32] IEEE 754 binary32 single-precision floating-point format.
\item[FP64] IEEE 754 binary64 double-precision floating-point format.
\item[GQA] grouped-query attention.
\item[GPTQ] post-training weight-quantization method.
\item[GSM8K] Grade School Math 8K benchmark.
\item[JSON] JavaScript Object Notation; used for structured evidence records.
\item[KL] Kullback--Leibler divergence.
\item[KV] key--value, typically attention key/value states.
\item[LLM] large language model.
\item[LoRA] low-rank adaptation.
\item[LZ / LZ77] Lempel--Ziv / Lempel--Ziv 1977 dictionary compression.
\item[MB / KB] megabyte / kilobyte; prose uses decimal byte units unless stated otherwise.
\item[MLP] multilayer perceptron.
\item[MSE] mean squared error.
\item[NLL] negative log-likelihood.
\item[PEFT] parameter-efficient fine-tuning.
\item[PPL] perplexity.
\item[PQF] Permute, Quantize, and Fine-Tune.
\item[QD] project shorthand for the orthogonal--diagonal alignment family used in the LoRA gauge probes.
\item[RD] rate--distortion.
\item[RFC] Request for Comments; RFC~1951 and RFC~3284 specify DEFLATE and VCDIFF, respectively.
\item[SiLU] sigmoid linear unit activation.
\item[SVD] singular value decomposition.
\item[SVD-LLM] singular-value-decomposition-based LLM compression method cited in Related Work.
\item[SwiGLU] Swish-gated linear unit feed-forward block.
\item[TV] total variation distance.
\item[VCDIFF] delta-encoding format defined by RFC~3284.
\item[ZipNN] lossless tensor/model compression library; ``NN'' denotes neural networks.
\item[Zstd] Zstandard lossless compression.
\end{description}

\paragraph{Mathematical and reporting notation.}
\begin{description}
\setlength{\itemsep}{0.15em}
\item[$\ell_1$ / $\ell_2$] entrywise absolute-value norm / Euclidean norm. The $\ell_1$ term generates soft-threshold sparsification proposals.
\item[$GL(r)$] general linear group of invertible $r\times r$ matrices.
\item[$Q_b$] significand-snap proposal at depth $b$; $b$ counts low-order significand-bit positions dropped when defining the coarser rounding grid, not deployment bitwidth.
\item[$L_s$] $\ell_1$ soft-threshold proposal tuned to target sparsity $s$.
\item[P99 / P99.9] 99th / 99.9th percentile of the reported token-wise statistic.
\item[acc / acc\_norm] raw accuracy / the \texttt{lm-eval} normalized-accuracy metric used for multiple-choice tasks.
\end{description}
\endgroup

\subsection{Datasets and evaluation corpora}
\begin{description}
\item[WikiText-2 raw (\texttt{Salesforce/wikitext:wikitext-2-raw-v1}) \citep{merity2017pointer}.]
Primary language-model corpus. Train is used for calibration/search, validation for independent held-out RD measurement and endpoint auditing, and test for final held-out audits. The raw variant retains original case, punctuation, and numbers. \emph{License/terms:} current Hub repository metadata lists \texttt{cc-by-sa-3.0} and the GNU Free Documentation License (GFDL). The card prose contains inconsistent CC BY-SA version wording; we therefore report the repository metadata rather than silently resolving the discrepancy.

\item[C4 English validation \citep{raffel2020t5}.]
Cross-domain language-model audit only. We stream the English validation split until the fixed token budget is assembled; C4 never selects candidate updates. \emph{License/terms:} Open Data Commons Attribution (ODC-BY) according to the current dataset card; users are additionally bound by Common Crawl terms for the underlying content.

\item[GSM8K \citep{cobbe2021gsm8k}.]
Grade-school mathematical-reasoning benchmark. In this paper it appears through the public Qwen3-GSM8K LoRA trajectory used in a negative/module-history study rather than as a main SmolLM2 evaluation set. \emph{License:} MIT in the original OpenAI \texttt{grade-school-math} repository.

\item[LAMBADA / \texttt{lambada\_openai} \citep{paperno2016lambada}.]
Post hoc next-word prediction benchmark evaluated only after compression endpoints are frozen. We use the standard \texttt{lm-eval} task implementation. \emph{License/terms:} the current \texttt{EleutherAI/lambada\_openai} Hub repository reports MIT metadata and describes the dataset license as Modified MIT; upstream terms remain authoritative.

\item[HellaSwag \citep{zellers2019hellaswag}.]
Post hoc commonsense sentence-completion benchmark included in the complete downstream-preservation audit. \emph{License:} MIT according to the current upstream/Hugging Face dataset card.

\item[PIQA \citep{bisk2020piqa}.]
Post hoc physical-commonsense multiple-choice benchmark. \emph{License/terms:} the current \texttt{ybisk/piqa} Hub card marks the dataset license as unknown; we therefore do not infer a dataset license from loader-code licensing or third-party mirrors.

\item[AI2 ARC Easy/Challenge \citep{clark2018arc}.]
Post hoc grade-school science multiple-choice benchmarks. ARC-Easy appears in the compact main-body preservation table; both Easy and Challenge are retained in the complete Appendix audit. \emph{License:} CC BY-SA 4.0 in the current \texttt{allenai/ai2\_arc} Hub metadata.
\end{description}

\subsection{Pretrained models and public adapters}
\begin{description}
\item[SmolLM2-135M (\texttt{HuggingFaceTB/SmolLM2-135M}) \citep{allal2025smollm2}.]
Principal dense causal language model for exact-gauge, activation-aware, output-functional, cross-codec, and held-out audits. Experiments operate on the public base checkpoint, not an instruction-tuned derivative. \emph{License:} Apache-2.0 on the Hugging Face model card/repository.

\item[SmolLM2-360M (\texttt{HuggingFaceTB/SmolLM2-360M}) \citep{allal2025smollm2}.]
Matched larger model for scale validation and for the 1-bit+$\ell_1$ and multibit+$\ell_1$ functional rate--distortion studies. \emph{License:} Apache-2.0 on the Hugging Face model card/repository.

\item[Qwen3-0.6B-Base (\texttt{Qwen/Qwen3-0.6B-Base}) \citep{yang2025qwen3}.]
Public Qwen3 base checkpoint used in module-symmetry controls, external-trajectory validation, and the dense architecture-transfer rate--distortion study. The repository revision is pinned when exact provenance is required. \emph{License:} Apache-2.0 on the Qwen model repository.

\item[Qwen3-GSM8K LoRA (\texttt{Neural-Hacker/Qwen3-GSM8K}).]
Rank-8 LoRA adaptation of Qwen3-0.6B used as one public base-to-adapter trajectory in the negative exact-gauge study. Its current card metadata is more restrictive than the Qwen3 base license. \emph{License:} CC BY-NC-SA 4.0 in current Hub metadata.

\item[Qwentaur-0.6B-LoRA-r8 (\texttt{socius/Qwentaur-0.6B-LoRA-r8}).]
Public rank-8 LoRA trajectory with checkpoints 500, 1000, 1500, 1878, and final; used to test exact temporal rank-gauge compression. \emph{License:} Apache-2.0 on the current Hub model card/repository.
\end{description}

\subsection{Software, serialization, and compression backends}
\begin{description}
\item[Python 3.] Experiment orchestration and reference implementation language. The standard-library \texttt{gzip} module is used for deterministic gzip streams in several paths. \emph{License:} Python Software Foundation License v2; incorporated components may carry additional licenses.

\item[PyTorch.] Tensor operations, model execution, autograd-based proposals, and behavioral audits. \emph{License:} BSD-style license in the PyTorch repository.

\item[NumPy.] Array utilities and numerical support used throughout the codebase. \emph{License:} BSD-3-Clause for the main project; binary distributions may include separately licensed bundled libraries.

\item[Hugging Face Transformers.] Public checkpoint/tokenizer I/O and evaluation; \emph{Apache-2.0}.

\item[Hugging Face Datasets.] Versioned loading and streaming of WikiText and C4. \emph{License:} Apache-2.0 for the library; individual datasets retain their own licenses above.

\item[Hugging Face PEFT.] Public LoRA-adapter loading. \emph{License:} Apache-2.0.

\item[EleutherAI \texttt{lm-eval}.] Zero-shot post hoc harness for LAMBADA, HellaSwag, PIQA, ARC-Easy, and ARC-Challenge. Recorded logs use task version 1, BF16 inference, and auto-detected batch size 64. \emph{License:} MIT.

\item[Safetensors.] Deterministic tensor checkpoint interchange and the canonical path used for ZipNN post hoc evaluation. \emph{License:} Apache-2.0.

\item[Zstandard / \texttt{zstd}.] Primary generic compressor; each byte-level experiment records the exact CLI/library version. \emph{License:} dual BSD or GPL-2.0.

\item[Brotli.] Generic cross-codec control and differentiable-surrogate mechanism study. \emph{License:} MIT.

\item[gzip / DEFLATE.] Generic cross-codec control with deterministic Python gzip framing. \emph{License:} Python Software Foundation terms for the wrapper; zlib license for the underlying implementation.

\item[ZipNN 0.5.4.] Tensor-specialized post hoc lossless codec on canonical safetensors; it never selects the main Zstandard endpoints. \emph{License:} MIT.
\end{description}

\paragraph{License scope.}
A dataset license does not determine a trained model's license, and a model card does not replace upstream data terms. This source/license reference is not legal advice or a redistribution grant; upstream licenses, notices, revisions, and terms remain canonical.

\section{Detailed Related Work}
\label{app:related}

\subsection{Functional non-identifiability and neural-network symmetries}
Many neural parameterizations are non-identifiable: multiple parameter vectors encode the same function. Positive-homogeneity symmetries in ReLU networks, for example, can change parameter-space sharpness without changing predictions, demonstrating that properties of a particular coordinate representation need not be properties of the represented function \citep{dinh2017sharp}. Our work takes a deployment-oriented view of the same phenomenon: if equivalent coordinates are physically different byte strings, the equivalence class itself becomes an optimization domain.

Permutation symmetry has received particular attention. \citet{entezari2022permutation} showed empirically that accounting for hidden-unit permutations can eliminate interpolation barriers between independently trained networks. \citet{ainsworth2023gitrebasin} developed practical permutation-alignment algorithms for merging independently trained models and characterized the resulting mode connectivity. Transformer-specific work has established broader permutation-equivariance properties across architectures such as BERT, GPT, and vision transformers \citep{xu2024permutation}. Particularly relevant is Permute, Quantize, and Fine-Tune (PQF) \citep{martinez2021permute}, which exploits adjacent-layer permutation invariance to search for functionally equivalent weight arrangements that are easier to vector-quantize and connects that search to rate--distortion theory. PQF is therefore an important precursor to the idea that symmetry can be selected for compressibility. Our setting differs in three respects: the objective is the measured output of an unchanged, potentially non-differentiable physical backend rather than a fixed product-quantization representation; the exact search includes scaling and basis gauges in addition to permutations; and the feasible set is then enlarged to empirical functional neighborhoods whose behavioral distortion is explicitly audited. These distinctions let us ask whether representation shaping transfers across independent lossless codecs and whether activation geometry can expose useful freedom beyond the exact symmetry orbit.

Model soups and related weight-space merging methods \citep{wortsman2022modelsoups} show that pretrained or fine-tuned solutions can often occupy compatible low-loss regions. This is conceptually adjacent to our approximate functional neighborhoods, but the goal differs: model merging combines information from separately optimized predictors, while our search starts from one model and explicitly constrains behavioral drift while optimizing physical representation cost.

Low-rank adaptation introduces another exact change-of-basis freedom. LoRA represents an update as a product of two low-rank factors \citep{hu2022lora}; the factorization is invariant to an invertible transformation in the latent rank space. We exploit this gauge when studying static adapters and checkpoint histories. The broader lesson is that factorized representations add coordinate freedoms beyond those already present in the dense network.

\subsection{Pruning, quantization, and low-rank compression}
Neural-network compression has a long history. Optimal Brain Damage and Optimal Brain Surgeon used curvature information to remove parameters while controlling loss \citep{lecun1989obd,hassibi1993obs}; later work combined pruning, quantization, and entropy coding into large end-to-end storage reductions \citep{han2016deepcompression}. The lottery-ticket literature further emphasized the existence of sparse subnetworks with favorable optimization properties \citep{frankle2019lottery}.

For modern language models, post-training methods aim to avoid the cost of full retraining. GPTQ uses approximate second-order information for accurate low-bit quantization \citep{frantar2023gptq}, while LLM.int8() isolates activation outliers to retain high-precision behavior where needed \citep{dettmers2022llmint8}. SparseGPT performs large-scale one-shot pruning through approximate sparse regression \citep{frantar2023sparsegpt}; Wanda provides a much simpler activation-weighted pruning criterion based on weight magnitudes and input activation norms \citep{sun2024wanda}. These methods alter sparsity or numerical representation to reduce inference/storage cost.

Activation-aware quantization and low-rank decomposition are especially relevant to our functional metric. AWQ identifies salient weights through observed activations and searches channel scalings that protect them during quantization \citep{lin2024awq}. ASVD argues that low-rank error should be measured through $\|(W-W')X\|_F$, not only through $\|W-W'\|_F$, and transforms weights according to activation statistics before truncation \citep{yuan2023asvd}. SVD-LLM similarly uses data-aware whitening and truncation-aware updates to better align low-rank approximation with model behavior \citep{wang2025svdllm}. Our local distortion in Eq.~\ref{eq:app-local} shares this output-space logic. The key difference is the controlled variable: we do not impose a target bitwidth, sparsity, or rank. Tensor shapes and dtypes remain unchanged, while candidate weights are selected according to the byte count produced by a separate lossless codec.

This distinction makes the approaches complementary. Quantization, pruning, and low-rank factorization seek new computational representations that directly reduce parameter count, precision, or runtime. Physical representation optimization can in principle be applied before or after such transformations whenever the resulting representation still contains exact or approximate functional degrees of freedom. Conversely, activation-aware compression supplies useful functional metrics for defining the admissible neighborhood in which a physical objective may be optimized.

\subsection{Lossless model compression and entropy coding}
Traditional compressors such as Zstandard, gzip/DEFLATE, and Brotli operate on byte streams without knowledge of neural semantics. Neural tensors have additional structure, motivating specialized lossless preprocessing and coding. ZipNN reorganizes floating-point neural weights according to dtype and numerical structure before lossless compression, reporting substantial storage and transfer savings for foundation models \citep{hershcovitch2025zipnn}. This line of work asks how a codec should exploit a fixed model representation.

Our question is complementary: given a fixed codec, can the \emph{model} move to a functionally equivalent or near-equivalent representation that the codec prefers? Cross-codec tests distinguish codec-specific effects from general representation shaping. Exact gauges transfer weakly between independently optimized codecs, consistent with codec-sensitive byte rearrangement, whereas activation-aware solutions transfer strongly across Zstandard, gzip, and Brotli. Tensor-aware ZipNN strengthens this complementarity result: Zstandard-selected 135M and 360M multibit+$\ell_1$ frontiers retain closely matched or slightly larger relative gains without ZipNN-driven search.

\subsection{Rate--distortion perspective}
Relaxing exact equality yields a model-specific rate--distortion problem: representation rate is traded against a fidelity criterion \citep{shannon1959rate}. Neural-weight compression has used rate--distortion ideas before, including permutation-assisted product quantization \citep{martinez2021permute}; here the rate variable is the measured output of the deployed physical backend:
\begin{equation}
\min_{\theta'} C_c(\theta')
\quad\text{s.t.}\quad
D_{\mathcal D}(f_{\theta'},f_\theta)\le\epsilon.
\end{equation}
Unlike conventional transform coding, however, the ``rate'' is the measured output of an arbitrary deployment codec and the distortion can be defined at multiple semantic levels: weight MSE, local activation reconstruction, block outputs, token-distribution KL, NLL, or downstream task metrics. Our experiments deliberately compare these nested notions rather than treating parameter distance as synonymous with functional distance. The empirical frontier therefore serves both as a compression result and as evidence about which local metrics best track end-to-end model behavior.

\end{document}